\pdfoutput=1
\documentclass[11pt]{article}

\usepackage[preprint]{acl}

\usepackage{times}
\usepackage{latexsym}
\usepackage{natbib}
\usepackage{amsmath}
\usepackage[T1]{fontenc}
\usepackage{amsfonts}
\usepackage{algorithm}
\usepackage{algorithmicx}
\usepackage{amssymb}
\usepackage{algorithm}
\usepackage{algpseudocode}
\usepackage{float}
\usepackage{stfloats}
\usepackage{subcaption}  % 子图与子标题
\usepackage{xcolor}
\usepackage{tikz} 
\usepackage[normalem]{ulem}
\usepackage{soul}
\usepackage{paralist}
\usepackage[most]{tcolorbox}
\usepackage{tcolorbox}
\usepackage{listings}
\usepackage{multirow}
\usepackage[utf8]{inputenc}

\usepackage{microtype}

\usepackage{inconsolata}

\usepackage{graphicx}

\title{From Data-Centric to Sample-Centric: Enhancing LLM Reasoning via Progressive Optimization}
\author{
  Xinjie~Chen$^{1,2}$ ,
  Minpeng~Liao$^{2,}$\thanks{\,\,Corresponding author. Work in progress.}\ ,
  Guoxin~Chen$^{2}$,
  Chengxi~Li$^{2}$,
  Biao~Fu$^{2}$,
  Kai~Fan$^{2,}$\footnotemark[1]\ ,
  Xinggao~Liu$^{1,}$\footnotemark[1] \\
  $^{1}$Zhejiang University \quad
  $^{2}$Alibaba Group Tongyi Lab \\
\texttt{\{xinjiechen, lxg\}@zju.edu.cn} \\ 
\texttt{\{minpeng.lmp, chenguoxin.cgx, xiji.lcx, fubiaobiao.fu, k.fan\}@alibaba-inc.com}
}
\begin{document}
\maketitle
\begin{abstract}
% Reinforcement learning with verifiable rewards (RLVR) has recently advanced the reasoning capabilities of large language models (LLMs). 
% While prior work has emphasized algorithmic design, data curation, and reward shaping, we investigate RLVR from a sample-centric perspective and introduce progressive optimization techniques. Our work addresses a critical question: how to best leverage a small set of trusted, high-quality demonstrations, rather than simply scaling up data volume.
% First, motivated by how hints aid human problem-solving, we propose \textbf{prefix-guided sampling}, an online data augmentation method that incorporates partial solution prefixes from expert demonstrations to guide the policy, particularly for challenging instances. 
% Second, inspired by how humans focus on important questions aligned with their current capabilities, we introduce \textbf{learning-progress weighting}, a dynamic strategy that adjusts each training sample's influence based on model progression. 
% We estimate sample-level learning progress via an exponential moving average of per-sample pass rates, promoting samples that foster learning and de-emphasizing stagnant ones. 
% Experiments on mathematical-reasoning benchmarks demonstrate that our methods outperform strong baselines, yielding faster convergence and a higher performance ceiling.
Reinforcement learning with verifiable rewards (RLVR) has recently advanced the reasoning capabilities of large language models (LLMs).
While prior work has emphasized algorithmic design, data curation, and reward shaping, we investigate RLVR from a sample-centric perspective and introduce \textbf{LPPO} (\textbf{L}earning-\textbf{P}rogress and \textbf{P}refix-guided \textbf{O}ptimization), a framework of progressive optimization techniques.
Our work addresses a critical question: how to best leverage a small set of trusted, high-quality demonstrations, rather than simply scaling up data volume.
First, motivated by how hints aid human problem-solving, we propose \textbf{prefix-guided sampling}, an online data augmentation method that incorporates partial solution prefixes from expert demonstrations to guide the policy, particularly for challenging instances.
Second, inspired by how humans focus on important questions aligned with their current capabilities, we introduce \textbf{learning-progress weighting}, a dynamic strategy that adjusts each training sample's influence based on model progression.
We estimate sample-level learning progress via an exponential moving average of per-sample pass rates, promoting samples that foster learning and de-emphasizing stagnant ones.
Experiments on mathematical-reasoning benchmarks demonstrate that our methods outperform strong baselines, yielding faster convergence and a higher performance ceiling.

\end{abstract}

% By leveraging a small amount of SFT data and exploration across temporal scales, our methods provide a new perspective on optimizing RLVR.
\section{Introduction}
Large Language Models (LLMs) have achieved significant advancements in complex reasoning, largely attributed to the paradigm of Reinforcement Learning with Verifiable Reward (RLVR)~\cite{1.3_RLVR, 1.3_RLVR, 1.1_deepseek_r1, 1.2_kimi_k1.5}. 
RLVR employs verifiable rewards to effectively guide solution exploration, and its potential was notably highlighted when Deepseek-R1-zero~\cite{1.1_deepseek_r1} demonstrated a pathway to enhance LLM reasoning via RLVR without necessitating supervised fine-tuning (SFT). 
This has spurred a considerable volume of research focused on advancing RLVR methods. 
Current efforts focus primarily on data curation~\cite{1.13_data_orz,1.14_data_bigmath, 1.15_data_limr, 1.16_data_limo}, reward design~\cite{1.13_data_orz,1.9_UnderstandingR1-Zero-LikeTraining_drgrpo, 1.17_data_logicRL, 1.18_reward_l1}, or refinement of core RL algorithms~\cite{1.9_UnderstandingR1-Zero-LikeTraining_drgrpo, 1.10_dapo, 1.11_RLVR, 1.12_REINFORCE++} from foundational RL methods like PPO \cite{1.5_ppo}, GRPO \cite{1.6_grpo}, and REINFORCE \cite{1.8_REINFORCE}. 
For example, \citet{1.13_data_orz} and \citet{1.10_dapo} filter out samples with either excessively high or zero pass rates based on the roll-out results of each problem. 
\citet{1.15_data_limr} proposes a learning impact measurement to select more valuable samples for training. 
\citet{1.10_dapo} and \citet{1.9_UnderstandingR1-Zero-LikeTraining_drgrpo} make improvements on the biased optimization of sequence length in GRPO. 
Meanwhile, \citet{1.19_vc_ppo} addresses issues in PPO via value pretraining and decoupled-GAE.

Despite recent advances, most existing methods either treat all training samples uniformly or rely on static heuristics, missing the opportunity to further exploit the potential of individual samples. While acquiring additional data can improve performance, data collection is often expensive or impractical. In such cases, it becomes critical to maximize the learning contribution from each available sample.

In contrast, human learners naturally focus more on challenging problems, and when faced with particularly difficult tasks, they seek hints or guidance from teachers or textbooks to acquire new techniques. As high-quality reasoning data become increasingly scarce, fully utilizing every available training instance is more important than ever. Thus, the central question shifts from “How can we collect more data?” to “How can we best leverage a small set of trusted solutions, especially when the policy gets stuck?”

To address this, we extend our focus from a purely data-centric paradigm to also include a sample-centric perspective. Specifically, we propose a sample-centric approach to reinforcement learning that aims to make the most of each solution-annotated sample, dynamically adjusting the optimization focus throughout training according to each sample’s learning trajectory.

First, to better guide exploration on difficult problems, we propose \textbf{Prefix-Guided Sampling (PG-Sampling)}—a data augmentation technique that uses partial solution prefixes from expert models. 
Unlike supervised learning or behavior cloning, PG-Sampling mimics the process of learning from a hint rather than from complete solutions. Inspired by human learning, this approach allows the model to benefit from high-quality guidance while preserving the exploratory advantages of reinforcement learning, striking a balance between the limitations of supervised fine-tuning and the instability of pure RL.

Second, also motivated by by the human learning process—where attention is often directed toward important questions in accordance to their current capabilities, we introduce \textbf{Learning-Progress Weighting (LP-Weighting)}, which dynamically adjusts each sample’s influence based on the dynamics of RL training progress. Unlike uniform or static weighting, LP-Weighting tracks per-sample progress and prioritizes those where the model is improving, allocating resources more efficiently and accelerating convergence.

Our contributions can be summarized as follows:
\begin{itemize}\itemsep0em
\item Inspired by the human learning process, we propose \textbf{LPPO} (\textbf{L}earning-\textbf{P}rogress and \textbf{P}refix-guided \textbf{O}ptimization), a novel sample-centric framework for RLVR that combines two complementary strategies: PG-Sampling and LP-Weighting.
\item We demonstrate through comprehensive experiments on reasoning benchmarks that our approach, LPPO, outperforms strong RLVR baselines.
\item We empirically find that our approach can achieve faster convergence and better generalization on both in-domain and out-of-domain problems.
\end{itemize}

\section{Related Work}
%需要介绍之前组内的工作，可以放在使用前缀/mcts等方法的介绍部分

\subsection{RLVR in LLMs}
Reinforcement Learning with Verifiable Reward (RLVR), where the reward is computed by a rule-based verification function, has been shown to be effective in improving the reasoning capabilities of LLMs. The most common practice of RLVR when applying reinforcement learning to LLMs on mathematical reasoning datasets is to use answer matching: the reward function outputs a binary signal based on whether the model’s answer matches the gold reference answer~\cite{2.1_rlvr,1.1_deepseek_r1,1.2_kimi_k1.5,2.2_rlvr,2.3_rlvr_cur,2.4_rlvr_cur_FastCuRL}. 
This reward design obviates the need for complex outcome-based or process-based reward models, offering a straightforward yet potent approach. 
The efficacy of RLVR is further bolstered by algorithmic advancements in reinforcement learning. These include optimizations to value functions or policy updates within PPO~\cite{1.5_ppo} (e.g., VinePPO~\cite{2.5_2_vineppo}, VCPPO~\cite{1.19_vc_ppo}, VAPO~\cite{2.5_vapo_rlvr}), methods for stabilizing and accelerating GRPO~\cite{1.6_grpo} (e.g., DAPO~\cite{1.10_dapo}, Dr. GRPO~\cite{1.9_UnderstandingR1-Zero-LikeTraining_drgrpo}, SRPO~\cite{1.7_SRPO}), and the integration of diverse algorithmic components (e.g. REINFORCE++~\cite{1.12_REINFORCE++}). 
Unlike these efforts, we focus on the temporal learning dynamics of individual samples within RLVR, which is an overlooked aspect. 

\subsection{Data Curation for LLM Post-Training}
Data curation for LLM post-training has been extensively studied~\cite{2.6_data}, with a significant focus on strategies for supervised fine-tuning (SFT), also known as instruction tuning. 
These strategies encompass LLM-based quality assessments~\cite{2.7_data_AlpaGasus}, leveraging features derived from model computations~\cite{2.8_data}, gradient-based selection techniques~\cite{2.9_data}, expert iteration based on Monte Carlo methods \cite{3_1.1_supermario,3_1.2_mcts_sft}, and knowledge distillation from expert model~\cite{1.16_data_limo}. 
Another line of research~\cite{2.10_data,2.10_2_data} investigates data selection for human preference datasets within the Reinforcement Learning from Human Feedback (RLHF) paradigm~\cite{2.11_data}.
While data curation is well-established for SFT and RLHF, strategies specifically for RLVR are comparatively less explored. One notable attempt is LIMR~\cite{1.15_data_limr}, which selected 1.4k examples from an 8.5k set for RLVR to match the performance of using the full set. 
Meanwhile, other studies~\cite{2.14_4sample, 2_3.5_one_sample} show that RLVR with very few examples can still improve performance, though with slower convergence. 
While insightful, most of these approaches focus on static data selection. More dynamic strategies, such as reverse curriculum reinforcement learning~\cite{r3_add}, offer a form of pre-set guided exploration, but leave online, sample-centric data augmentation for challenging samples largely unaddressed.
% Unlike prior RLVR data curation methods that emphasize selection, our Prefix-Guided Sampling focuses on augmenting and guiding difficult cases using partial solutions, enhancing RL exploration and balancing generalization with convergence.

\section{Methodology}
\label{sec:methods}
\subsection{From Data-Centric to Sample-Centric}
\label{subsec:From Data-Centric to Sample-Centric RLVR}

The scaling law of data plays a crucial role in LLM training~\cite{3_1.5_scaling_law, 3_1.6_scaling_law}. 
During the post-training phase of LLM, data-centric strategies are critical to enhance efficiency and performance~\cite{3_1.4_survey}. 
However, high-quality reasoning data, particularly those involving complex reasoning are costly to acquire and often scarce. 
Consequently, a sole reliance on data-centric strategies typically encounters diminishing marginal returns.

This challenge underscores the necessity of shifting the focus towards a ``\textbf{sample-centric}'' strategy. 
The core objective of this strategy is, given a high-quality dataset, to efficiently utilize each training sample by optimizing the presentation order of samples, selection criteria, or their contribution to model updates during training, thereby maximizing learning efficacy. 
Existing machine learning techniques such as curriculum learning~\cite{3_1.7_curriculum, 3_1.8_curriculum}, active learning~\cite{3_1.8_active_learning}, hard example mining \cite{3_1.10_hard_example_mining}, and sample weighting \cite{3_1.9_sample_weighting} all exemplify this approach. 
Considering the scarcity of high-quality reasoning data and the imperative to maximize the utility of existing data resources, we posit that deeper exploration at the sample-centric level is crucial, especially for complex reasoning tasks.

Following this principle, this paper, within the RLVR framework and from a sample-centric perspective, proposes two specific methods: (1) \textbf{Prefix-Guided Sampling}, a data augmentation strategy that utilizes partial expert solution prefixes to guide model exploration on challenging samples; and (2) \textbf{Learning-Progress Weighting}, a mechanism that adjusts the training influence of samples based on the dynamic changes in the model's learning progress on each sample.

\subsection{Prefix-Guided Sampling}
\label{subsec:pg_sampling}
To better guide the exploration process, particularly for challenging problems, we introduce Prefix-Guided Sampling (PG-Sampling).
This data augmentation strategy is inspired by partially observable reasoning trajectories.
The technique involves guiding the policy by providing partial solutions as hints for difficult training samples.
These prefixes are sampled from successful solutions generated by expert models.

Unlike RLVR using problem-answer datasets $D=\{q,a\}$, PG-Sampling requires datasets $D_{pg}=\{q,S_{\text{exp},q},a\}$, where $S_{\text{exp},q}$ denotes the expert solution for a problem $q$. 
Thus, let $Q_{\text{sol}}$ be the set of problems for which such expert solutions $S_{\text{exp},q}$ are available.
From this set, a problem $q \in Q_{\text{sol}}$ is deemed ``challenging'' at a given training epoch $t$ if its pass rate falls at or below a predefined threshold $\epsilon_c$ \footnote{$\epsilon_c=0$ in our experiments, indicating that PG-Sampling is applied to problems the model fails to solve without guidance.
}.

For a challenging problem $q \in Q_{\text{sol}}$, a prefix $S_{\text{pre},q}$ is generated from its expert solution $S_{\text{exp},q} = (y_1, y_2, \dots, y_M)$, where $M$ is the total number of tokens in the expert solution.
The desired length of prefix, $L_p$, is determined by:
\begin{equation}
L_p = \lfloor \lambda \cdot M \rfloor
\label{eq:prefix_length}
\end{equation}
where $\lambda$ is a truncation ratio randomly sampled from a uniform distribution $\mathcal{U}(\beta_{\min}, \beta_{\max})$ \footnote{We set $\beta_{\min}=0.3, \beta_{\max}=0.8$ in our experiments.}, $\lfloor \cdot \rfloor$ means $L_p$ is further clipped to the last newline character to ensure the prefix ends at a complete line.
The prefix $S_{\text{pre},q}$ then consists of the first $L_p$ tokens of $S_{\text{exp},q}$:
\begin{equation}
S_{\text{pre},q} = (y_1, y_2, \dots, y_{L_p})
\label{eq:prefix_definition}
\end{equation}

The policy $\pi_{\theta}$ then generates the remainder of the solution $S_{\text{rem},q}$ as follows:
\begin{equation}
S_{\text{rem},q} \sim \pi_{\theta}(\cdot | q \oplus S_{\text{pre},q})
\label{eq:remainder_generation}
\end{equation}
where $\oplus$ denotes token sequence concatenation.
Specifically, the solution derived from $S_{\text{rem},q}$ is evaluated for the reward signal of RL training.
% The reward $r_q$ for this trajectory can then be determined by a verifier or reward function $V$, which compares $S_{remainder,q}$ with the ground truth answer $a$ from the dataset $D_{pg}$:
% %
% \begin{equation}
% r_q = V(\text{ExtractAnswer}(S_{\text{rem},q}), a)
% \label{eq:reward_calculation}
% \end{equation}
% %
% This reward $r_q$ is then used to update the policy $\pi_{\theta}$ using standard reinforcement learning algorithms.

% PG-Sampling allows the model to learn from high-quality solution structures while still requiring it to generate the remaining parts of the solution, thus maintaining diversity and leveraging the strengths of both RL (self-exploration) and SFT (learning from examples). 
% Unlike prior RLVR data curation methods that primarily focus on selection, PG-Sampling focuses on augmenting and guiding difficult cases using partial solutions.

PG-Sampling guides the model using partial expert solutions, encouraging it to complete the remaining and thus balancing self-exploration with learning from examples. 
Unlike prior RLVR data curation methods that focused only on selection, PG-Sampling emphasizes augmenting and guiding challenging cases. While our approach of using partial trajectories shares a surface-level similarity with curriculum-based methods like Reverse Curriculum Reinforcement Learning~\cite{r3_add}, our online, sample-centric triggering mechanism is fundamentally different. A detailed comparison is provided in Section 2.2.

\subsection{Learning-Progress Weighting}
\label{subsec:lp_weighting}

We introduce Learning-Progress Weighting (LP-Weighting), a method to dynamically re-weight each sample's \textit{advantage estimate} in reinforcement learning. 
This re-weighting is implemented by applying a dynamic scaling factor, which is itself determined by the model's learning progress on that sample—specifically, its improvement (\emph{e.g.}, pass rate) between adjacent epochs. 
The objective is to thereby amplify the influence of training samples where the model is actively improving and diminish that of samples where learning has stagnated or degraded.

For each sample (indexed by $i$), we track the Exponential Moving Average (EMA)~\cite{3_1.11_ema} of its pass rate at epoch $t$, denoted as $\overline{p_i}(t)$. 
The raw pass rate, $pass\_rate_i(t)$, derived from a finite number of rollouts, can exhibit high variance. 
The application of EMA helps to smooth these random fluctuations, providing a more stable assessment of the learning state:
\begin{equation}
\overline{p_i}(t) = \alpha \cdot pass\_rate_i(t) + (1-\alpha) \cdot \overline{p_i}(t-1)
\label{eq:ema_pass_rate}
\end{equation}
where $\alpha$ is the smoothing factor. 
The initial $\overline{p_i}(0)$ is set to the pass rate observed at the first epoch.

Based on the smoothed pass rate, the learning progress for sample $i$ at epoch $t$, denoted $\Delta_i(t)$, is defined as the first-order difference of its EMA pass rate:
\begin{equation}
\Delta_i(t) = \overline{p_i}(t) - \overline{p_i}(t-1)
\label{eq:learning_progress_delta}
\end{equation}

This learning progress $\Delta_i(t)$ is then used to compute a dynamic weight $w_i(t)$ for sample $i$:
\begin{equation}
w_i(t) = \sigma(\kappa \cdot \Delta_i(t)) + b
\label{eq:dynamic_weight}
\end{equation}
where $\sigma$ represents the sigmoid activation function, $\kappa$ and $b$ are factors that control the sensitivity and bias of the final weight to the learning progress.\footnote{We set $\kappa$=8.0 and $b$=0.5 in our experiments.} 
The dynamic weight $w_i(t)$ is used during the RL policy update phase to adjust the advantage estimate $\hat{A}_i$ for sample $i$. 
The weighted advantage, denoted as $\hat{A}'_i$, is calculated as:
\begin{equation}
\hat{A}'_i = w_i(t) \cdot \hat{A}_i
\label{eq:weighted_advantage}
\end{equation}
Alternatively, $w_i(t)$ can also be used as a sampling probability to determine if sample $i$ is selected for the current training batch.

The LP-weighting method can be easily integrated with Group Relative Policy Optimization (GRPO)~\cite{1.6_grpo}. 
%GRPO aims to reduce RL training costs by forgoing a critic model, instead estimating the baseline from group scores. 
The policy $\pi_\theta$ is optimized by maximizing the following objective:

{\small
\begin{equation}
\begin{split}
& \mathcal{J}_\text{LP-GRPO}(\theta)=\mathbb{E}\left[q_i \sim P(Q),\left\{o_{i,k}\right\}_{k=1}^{G} \sim \pi_{\theta_{\text {old }}}(O \mid q_i)\right] \\
& \left[ \frac{1}{G} \sum_{k=1}^G \min \left( \rho_{i,k}(\theta) \hat{A}'_{i,k}, \right. \right.
\left. \left. \text{clip}(\rho_{i,k}(\theta), 1-\epsilon, 1+\epsilon) \hat{A}'_{i,k} \right) \right]
\end{split}
\label{eq:lp_grpo_objective_no_kl}
\end{equation}
}
where $\rho_{i,k}(\theta) = \frac{\pi_\theta(o_{i,k}|q_i)}{\pi_{\theta_{old}}(o_{i,k}|q_i)}$ is the probability ratio for output $o_{i,k}$ of question $q_i$. 
For each question $q_i$ (drawn from a data distribution of questions $P(Q)$), it performs rollouts with a group of size $G$, $\{o_{i,k}\}_{k=1}^G$ from the old policy $\pi_{\theta_{old}}(O \mid q_i)$. To encourage broader exploration, the KL penalty term is omitted from the objective.

% To encourage broader exploration, the KL divergence loss is omitted from the objective. 

In our context, the question $q_i$ serves as the conditioning input to the policy $\pi_\theta$; thus, the policy is conditioned on the actual content of question $q_i$, analogous to $q$ in the original GRPO formulation. 
The term $\hat{A}'_{i,k}$ is the LP-weighted advantage for output $o_{i,k}$ of question $q_i$, calculated in the same way as Eq.~(\ref{eq:weighted_advantage}).
%
% \begin{equation}
% \hat{A}'_{i,k} = w_i(t) \cdot A_{i,k}
% \label{eq:lp_weighted_grpo_advantage}
% \end{equation}
%
%Here, $w_i(t)$ is the LP-weight for the sample corresponding to question $q_i$ (as in Eq. \ref{eq:lp_grpo_objective_no_kl}) for which the group of $G$ outputs $\{o_{i,k}\}_{k=1}^G$ is generated. 
%This weight is derived from Eq.\ref{eq:dynamic_weight} using the learning progress of this $i$-th sample. 
%$A_{i,k}$ is the original advantage term from GRPO for the $k$-th output $o_{i,k}$ of question $q_i$ in the group, computed as a group-normalized reward:
%
% \begin{equation}
% A_{i,k} = \frac{r_{i,k} - \text{mean}(\{r_{i,1}, r_{i,2}, \ldots, r_{i,G}\})}{\text{std}(\{r_{i,1}, r_{i,2}, \ldots, r_{i,G}\}) + \delta_{std}}
% \label{eq:grpo_base_advantage}
% \end{equation}
%
%where $\{r_{i,1}, r_{i,2}, \ldots, r_{i,G}\}$ are the actual rewards for the $G$ outputs in the group generated for question $q_i$, $r_{i,k}$ is the specific reward for the $k$-th output of question $q_i$, and $\delta_{std}$ is a small constant for numerical stability.

Intuitively, if $\Delta_i(t) < 0$, it indicates that the policy rollouts for the $i$-th problem at epoch $t$ perform worse than in the previous epoch. 
In such cases, we prefer not to emphasize these regressions during policy updates, avoiding reinforcement of  ``stagnant'' or ``degraded'' solutions. 
Conversely, this mechanism, samples with positive learning progress (\emph{i.e.}, $\Delta_i(t) > 0$) receive higher weights, correspondingly increasing their contribution to model parameter updates. 
To prevent the complete neglect of persistently challenging samples and to mitigate potential catastrophic forgetting, a minimum lower bound $b$ is set for the weights $w_i(t)$, which is indicated in Eq.~(\ref{eq:dynamic_weight}).

The LP-Weighting method aims to automatically direct the model's optimization focus towards samples where it is making substantial learning gains. 
This dynamic adjustment strategy is expected to enhance the overall efficiency and convergence performance of the training process, particularly when encountering training bottlenecks.

\subsection{Detailed Training Algorithm}
\label{subsec:integrated_training}

Integrating PG-Sampling and LP-Weighting into the RLVR training pipeline introduces several distinctions from the standard RLVR process. These are primarily centered around data preparation, batch construction, sample weighting, and online data curation:

\paragraph{Data Preparation} The training data is categorized into two types: standard RLVR data denoted as $D$, and specialized data designated for PG-Sampling, $D_{pg}$ (as detailed in Section \ref{subsec:pg_sampling}). The $D_{pg}$ dataset comprises challenging problems, each accompanied by an expert solution. It is also feasible to conduct training using only $D_{pg}$, in which case all problems processed during training are equipped with expert solutions.

\paragraph{Batch Construction with PG-Sampling}
At each training step, samples in the current batch undergo an initial rollout to calculate their pass rates. Samples with non-zero pass rates proceed with standard RL operations (e.g., advantage calculation, policy updates). Conversely, "difficult" samples from $D_{pg}$ (those with zero pass rate) are augmented with a partial expert solution prefix and then stored for inclusion in the new batch for re-evaluation during the subsequent training step. This ensures these prefix-augmented difficult samples are revisited to maximize learning.

\paragraph{LP-Weighting Calculation}
For each sample $i$ in a training batch, its pass rate $pass\_rate_i(t)$ is used to update its EMA pass rate $\overline{p_i}(t)$, subsequently determining the learning progress $\Delta_i(t)$ and the dynamic weight $w_i(t)$, as detailed in Section \ref{subsec:lp_weighting} (Eq. \ref{eq:ema_pass_rate} to \ref{eq:dynamic_weight}). During the policy update phase (e.g., using GRPO), these dynamic weights $w_i(t)$ are applied to the advantage values of their respective samples (Eq. \ref{eq:weighted_advantage}).

\paragraph{Online Data Curation}
% ($D_{active}$)
To maintain training efficiency and focus on more instructive examples, samples that achieve very high pass/low rates will be excluded from the current batch and the next training epoch \footnote{We exclude samples with a 100\% or 0\% pass rate during sampling, and exclude samples with a 100\% pass rate for the next epoch in our experiments.}. This adaptive data curation strategy allows the model to concentrate computational resources on more challenging or less frequently solved problems, thereby optimizing the learning process. We use this strategy in all our experiments, including the GRPO baseline.

The pseudo code of proposed algorithm is detailed in Appendix \ref{sec:Integrated Training Process with PG-Sampling and LP-Weighting}.

\section{Experiments}
\subsection{Experimental Setup}

\paragraph{Base Model} To ensure consistency with prior studies, we use Qwen2.5-Math-7B \cite{4_1_qwen_math}, a strong publicly available LLM with fundamental mathematical reasoning capabilities, as the base model for RL fine-tuning.

\paragraph{Datasets}
\textit{Training Data:}
Our training dataset is intentionally small and curated for quality. It comprises:
1. The dataset with expert solutions, denoted as $D_{pg}$ in Section \ref{subsec:pg_sampling}, consists of the same 817 examples as used in LIMO \cite{1.16_data_limo}. These examples are high-quality mathematical problems selected from sources such as NuminaMath \cite{4_11_Numinamath} and historical AIME problems (1983–2023) \cite{aime_1983_2023}. They were specifically chosen based on their difficulty, generality, and diversity in required mathematical knowledge, thereby designed to elicit complex reasoning and accompanied by detailed step-by-step solutions provided by expert models \cite{1.1_deepseek_r1, 4_2_o1_journey}.
2.  Level 3-5 problems from the MATH dataset \cite{4_3_math_dataset}, following the setup used in \citet{1.9_UnderstandingR1-Zero-LikeTraining_drgrpo}. These slightly easier problems establish foundational skills and ensure broader competency, preventing over-specialization on the hardest tasks.

\textit{Evaluation Benchmarks:} We evaluate on a diverse set of mathematical reasoning benchmarks: AIME24, AIME25, MATH-500 \cite{4_3_math_dataset, 4_4_math500_dataset}, AMC23 \cite{4_11_Numinamath}, MinervaMath \cite{4_4_Minerva_dataset}, and OlympiadBench \cite{4_5_OlympiadBench}.

\paragraph{Evaluation Metrics} Following the settings in \cite{1.1_deepseek_r1, 1.9_UnderstandingR1-Zero-LikeTraining_drgrpo, 4_6_GPG}, the primary evaluation metric is \textit{pass@1}, which is defined as the percentage of problems for which the model generates the correct final answer. To ensure fair comparisons, all reproduction results from third-party ~\cite{4_7_sober_repro_eval} or our own, which are tagged with $\dagger$ in Table \ref{tab:main_results}, are averaged over three runs (\textit{avg@3}). 

\paragraph{Implementation Details} We use the verl \cite{4_8_verl} pipeline for RL fine-tuning and evaluation. By default, the coefficients for entropy loss is set to -0.001. For training rollouts generated via vLLM \cite{4_9_vllm}, we set the sampling temperature to $1.0$. The optimizer uses a learning rate of $1 \times 10^{-6}$ and a weight decay coefficient of $0.01$. We use a training batch size of 128, a mini-batch size of 64, and generate 32 rollouts per sample. To accommodate potentially long sequences required by PG-Sampling, we set the maximum prompt lengths to 5000 tokens each, and the maximum response length is also 5000. Considering that the Qwen2.5-Math-7B model has only a 4096 context length by default, we expand this to 10,000. We store the model checkpoint every 5 steps for evaluation. Each experiment is trained for 300 steps on 8 A100 GPUs. The details of prompt and reward design are shown in Appendix \ref{sec:AdditionalExperimentSettings}.

\subsection{Main Results}
% 主实验部分。和其他所有从Qwen-math/base-7b进行RLVR训练的模型进行比较。
\begin{table*}[t]
\centering
\resizebox{\textwidth}{!}{    % 适配整页宽
\begin{tabular}{lcccccccc}
\hline
\textbf{Model (7B)}   & \textbf{AIME24} & \textbf{AIME25} & \textbf{AMC23} & \textbf{MATH-500} & \textbf{Minerva} & \textbf{OlympiadBench} & \textbf{Average} & \textbf{Samples (k)} \\ \hline
Qwen-2.5-Math-7B-Instruct $\dagger$         & 15.7  & 10.7        & 67.0        & 82.9        & 35.0 & 41.3        & 42.1 & --   \\
Qwen2.5-Math-7B (Base) $\dagger$            & 20.7  & 8.7         & 56.2        & 64.3        & 17.3 & 29.0        & 32.7 & --   \\
Eurus-2–7B-PRIME                            & 26.7  & --          & 57.8        & 79.2        & 38.6 & 42.1        & --   & 150 \\
Eurus-2–7B-PRIME $\dagger$                  & 17.8  & 14.0        & 63.0        & 80.1        & 37.5 & 43.9        & 42.7 & 150 \\
Oat-Zero-7B                                 & 43.3* & --          & 62.7        & 80.0        & 30.1 & 41.0        & --   & 8.5 \\
Oat-Zero-7B $\dagger$                       & 28.0  & 8.8         & 66.2        & 79.4        & 34.4 & 43.8        & 43.4 & 8.5 \\
OpenReasoner-Zero-7B                        & 17.9  & 15.6        & --          & 81.4        & --   & --          & --   & 129 \\
OpenReasoner-Zero-7B $\dagger$              & 19.7  & 15.7        & 59.5        & \textbf{83.9*} & 31.6 & \textbf{47.6*} & 43.0 & 129 \\
SimpleRL-Zero-MATH-7B                       & 24.0  & --          & 70.0        & 80.2        & 37.5 & 39.0        & --   & 8.5 \\
SimpleRL-Zero-MATH-7B $\dagger$             & 22.7  & 10.7        & 62.2        & 76.9        & 30.1 & 39.3        & 40.3 & 8.5 \\
GPG-7B                                      & 33.3  & --          & 65.0        & 80.0        & 34.2 & 42.4        & --   & 17  \\
GPG-7B $\dagger$                            & 26.7  & 10.0        & \textbf{75.0*} & 79.8 & 39.3 & 44.7        & 45.9 & 17  \\
LIMR-7B                                     & 32.5  & --          & 63.8        & 78.0        & --   & --          & --   & 1.4 \\
LIMR-7B $\dagger$                           & 30.7  & 7.8         & 62.2        & 76.5        & 34.9 & 39.3        & 41.9 & 1.4 \\ \hline
Qwen2.5-Math-7B + GRPO (Baseline)           & 30.0  & 10.0        & 62.5        & 80.7        & 39.4 & 43.1        & 44.3 & 9.2 \\
\quad + LP-Weighting (Ours)                 & 36.7  & \textbf{16.7*} & 64.1 & 81.0 & 39.6 & 42.9 & 46.8 & 9.2 \\
\quad + LP-Weighting + PG-Sampling (Our LPPO)   & \textbf{40.0} & \textbf{16.7*} & 69.2 & 82.0 & \textbf{41.5*} & 43.5 & \textbf{48.8*} & 9.2 \\ \hline
\end{tabular}
}
\caption{Zero-shot \textit{pass@1} performance of 7B models on six mathematical reasoning benchmarks. $\dagger$ indicates results obtained from reproductions. \textbf{Bolded} values denote the best reproducible results, and asterisked* values represent the best results among all reported results.}
\label{tab:main_results}
\end{table*}

\paragraph{Comparison Methods}
The proposed sample-centric methods are benchmarked against 7B models.
Our direct RLVR baseline applies GRPO \cite{1.6_grpo} to the Qwen2.5-Math-7B model \cite{4_1_qwen_math}.
We further compare our results with several contemporary 7B models fine-tuned using RLVR without additional supervised fine-tuning beyond their initial pre-training, including Eurus-2-7B-PRIME \cite{4_10_Eurus_prime}, Oat-Zero-7B \cite{1.9_UnderstandingR1-Zero-LikeTraining_drgrpo}, OpenReasoner-Zero-7B \cite{1.13_data_orz}, SimpleRL-Zero-MATH-7B \cite{4_9_simplerl}, GPG-7B \cite{4_6_GPG}, and LIMR-7B \cite{1.15_data_limr}. For broader context, the performance of the base Qwen2.5-Math-7B and its instruction-tuned version, Qwen-2.5-Math-7B-Instruct, are also presented. For a fair and comprehensive comparison, results for external methods sourced from reproductions and original publications/papers are both presented.

Table \ref{tab:main_results} presents the main performance comparison on zero-shot \textit{pass@1} across six mathematical reasoning benchmarks.
Compared to the direct GRPO baseline, applying LP-Weighting alone consistently improves performance across all benchmarks, lifting the average score to 46.8\% (+2.5\%). This demonstrates the general effectiveness of dynamically adjusting sample influence based on learning progress.
Integrating PG-Sampling on top of LP-Weighting yields further significant gains, achieving the highest average score of 48.8\% among all listed models. This combined approach substantially outperforms the GRPO baseline (+4.5\%) and also surpasses LP-Weighting alone (+2.0\%), highlighting the strong additive benefit derived from using prefix guidance, particularly on challenging problems (e.g., boosting AIME24 score from 30.0\% to 40.0\%).

Our best model, LP-Weighting + PG-Sampling, also achieves the highest average performance among all strong contemporary 7B models. Specifically, it achieves state-of-the-art results on AIME24, AIME25, and Minerva. While some specialized models show higher performance on individual benchmarks, our approach provides the best overall balance and generalization.

% In essence, our sample-centric strategies, LP-Weighting and PG-Sampling, achieving superior average performance compared to existing methods.

\newcommand{\legitem}[2]{%
  \tikz[baseline=0ex]{%
    \draw[#1, line width=1.3pt, line cap=round]
          (0,0.3em) -- (1.4em,0.3em);      % 斜线段
  }%
  \hspace{0.4em}#2%
}
\definecolor{lpgreen}{RGB}{214,238,202}
\definecolor{pgblue}{RGB}{88,182,205}
\definecolor{pgdeepblue}{RGB}{66,86,165}
\begin{figure*}[t]
  \centering
  % ==== 四张子图 =============================================================
  \begin{subfigure}[b]{0.245\textwidth}
    \includegraphics[width=\linewidth]{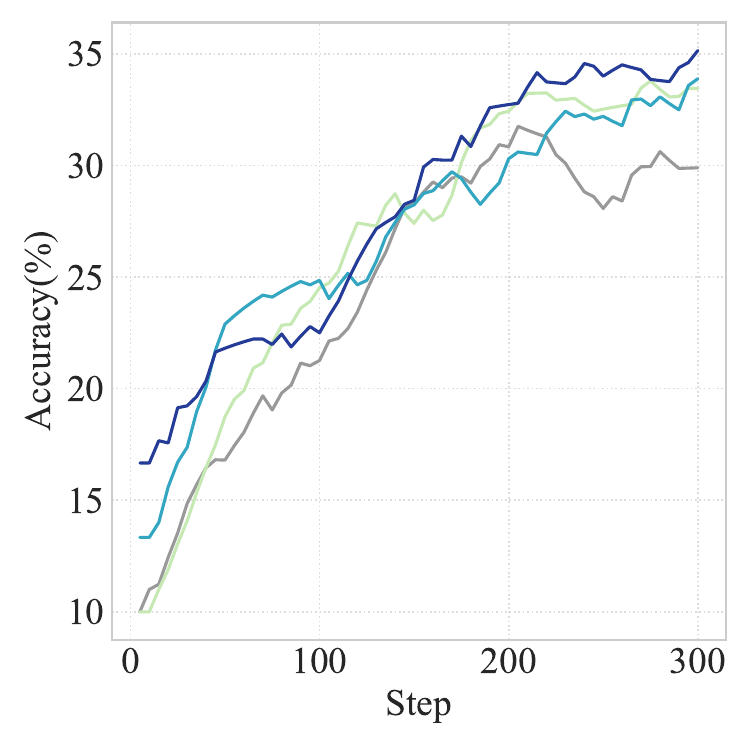}
    \caption{AIME2024}
    \label{fig:sub1}
  \end{subfigure}
  \hfill
  \begin{subfigure}[b]{0.245\textwidth}
    \includegraphics[width=\linewidth]{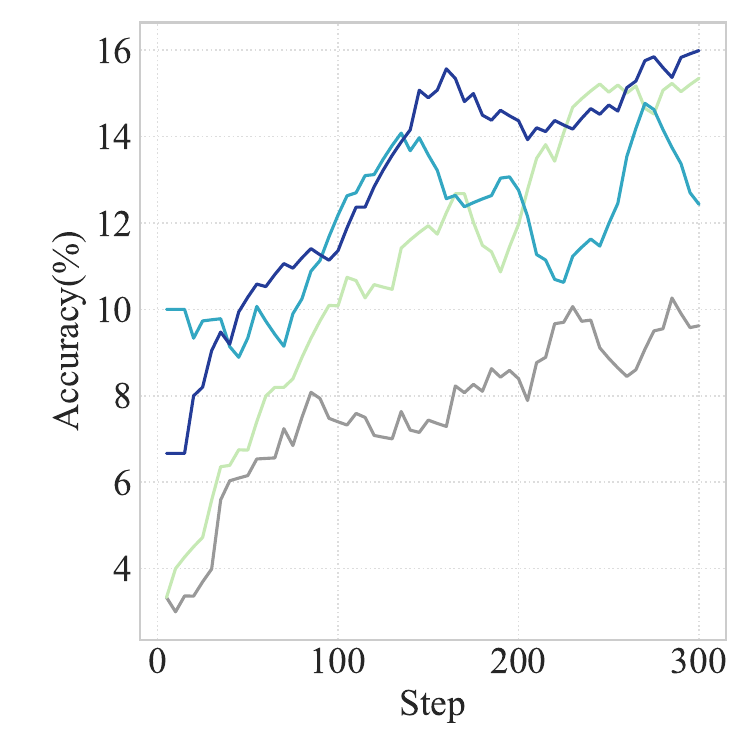}
    \caption{AIME2025}
    \label{fig:sub2}
  \end{subfigure}
  \hfill
  \begin{subfigure}[b]{0.245\textwidth}
    \includegraphics[width=\linewidth]{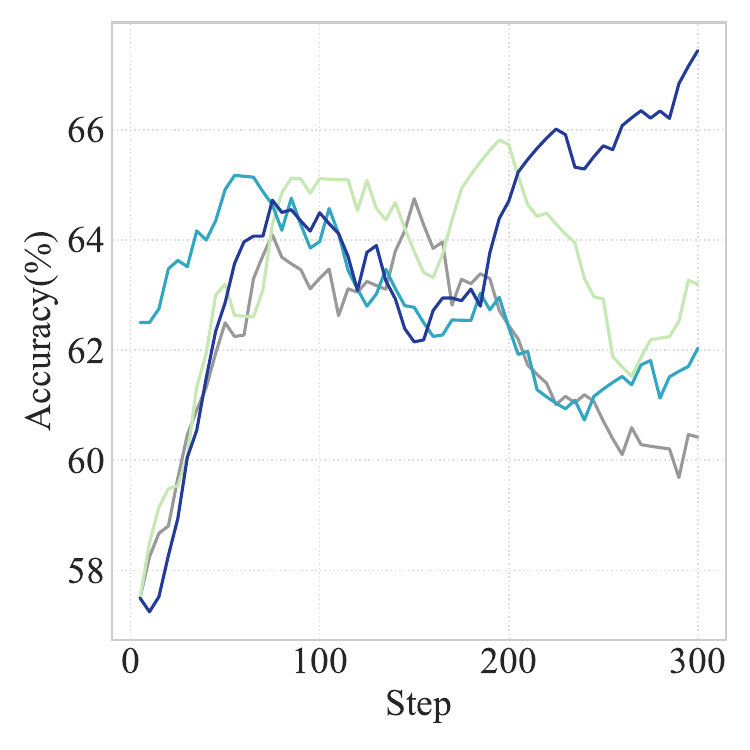}
    \caption{AMC23}
    \label{fig:sub3}
  \end{subfigure}
  \hfill
  \begin{subfigure}[b]{0.245\textwidth}
    \includegraphics[width=\linewidth]{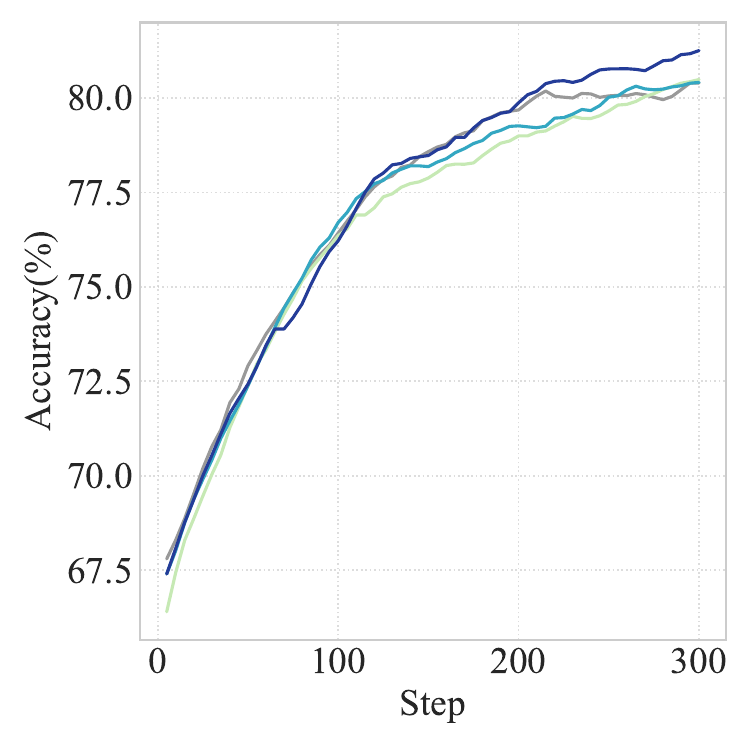}
    \caption{MATH500}
    \label{fig:sub4}
  \end{subfigure}

  % ======================= Legend ===========================================
  \vspace{0.2em} % 与子图稍作间隔
  \begin{minipage}{0.95\linewidth}
    \centering
    \legitem{gray!70}{\small GRPO}%
    \hspace{1em}% 调整不同图例项之间的水平间距
    \legitem{lpgreen}{\small GRPO + LP-Weighting}%
    \hspace{1em}
    \legitem{pgblue}{\small GRPO + PG-Sampling}%
    \hspace{1em}
    \legitem{pgdeepblue}{\small GRPO + PG-Sampling + LP-Weighting}
  \end{minipage}

  \caption{Ablation studies of our proposed LP‑Weighting and PG‑Sampling.}
  \label{fig:ablation}
\end{figure*}

\subsection{Ablation Studies} 
\label{subsec:ablation}

In Figure~\ref{fig:ablation}, we examine how different sample-centric methods influence the accuracy trajectories on four mathematical benchmarks throughout training. All curves are \mbox{EMA}-smoothed with $\alpha\!=\!0.9$ and raw fluctuations are therefore slightly underestimated.

\paragraph{Early-stage acceleration (\emph{Epoch~1})}
During the first epoch ($\le\!60$ steps), the two strategies that include PG-Sampling dominate the baseline on every benchmark, while the \textsc{LP-only} variant follows the baseline closely.  
Because LP-Weighting requires the pass rates from the previous epoch to compute dynamic advantages, its weights are still uniform at this stage; consequently, the optimization policies of \textsc{PG} and \textsc{PG+LP} are identical, and the visible oscillation stems solely from the randomly drawn reasoning prefixes and the training noise.

\paragraph{Mid-/late-stage gains (\emph{Epoch~2+})}
From step ~$\sim$60 onwards, the influence of LP-Weighting manifests as a steeper slope and lower variance: samples that achieve higher imporvement on pass rates are up-weighted, which filters noisy gradients and, once a hard question starts to yield improvement, quickly amplifies its influence.
The compound variant consequently outperforms all others, showing that the two techniques are complementary rather than redundant.

Building on above observations, we now examine how the two mechanisms steer learning dynamics: \textbf{Fast start.} PG-Sampling appends a solution prefix, giving an immediate accuracy jump and reducing exploration steps;  \textbf{Reliable finish.} from epoch 2 onward, LP-Weighting shifts attention toward samples that raise pass rates, filtering gradient noise and lifting the ceiling. Used jointly, the two mechanisms drive the quick convergence and high accuracy on every benchmark.

%阐明KL可能带来的性能限制和风险。对于baseline来说，曾观察到带KL时继续训练存在崩溃，但我们的方法并未观察到相关现象。
% 是否存在kl带来的影响/prefix采样比例给性能带来的影响等

\subsection{Experiment Analysis}

%针对prefix策略的分析

\begin{figure}[htbp]
    \centering
    \includegraphics[width=0.38\textwidth]{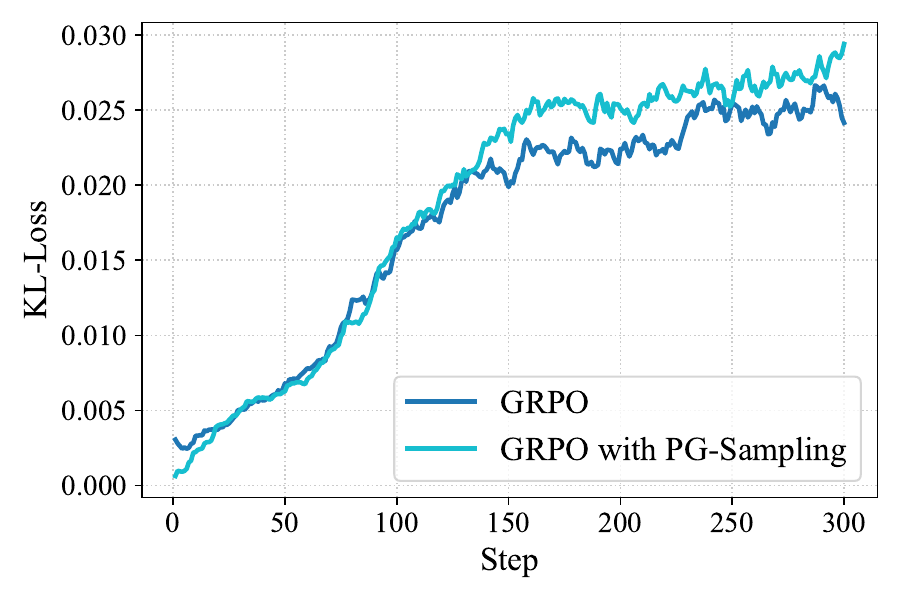} % 替换路径
    \caption{The KL Divergence with/without PG-Sampling (EMA smoothed with $\alpha$=0.9).}
    \label{fig:kl_Loss}
\end{figure}

\paragraph{KL Divergence}
 % We monitor the KL loss between the policy model and the reference model to assess exploration during training.
Although not optimized, the KL divergence between the policy and reference models is monitored to gauge exploration.
 As shown in Fig.~\ref{fig:kl_Loss}, the PG-Sampling variant diverges from the reference distribution more rapidly after around 150 updates, indicating that injecting a prefix after the question encourages greater exploration. After step 250, both curves reflect similar distances from the reference, but not necessarily similar policies. Combining the results in Fig.~\ref{fig:ablation} and Fig.~\ref{fig:kl_Loss}, we observe that although the PG-Sampling strategy does not further increase exploration divergence, its performance improves significantly. This indicates that PG-Sampling utilizes the exploration budget more efficiently and guides the policy toward higher-value regions of the solution space.

\begin{table*}[!ht]
\centering
\resizebox{\textwidth}{!}{
\begin{tabular}{lccccccc}
\hline
Model (7B) & AIME24 & AIME25 & AMC23 & MATH500 & Minerva & OlympiadBench & Average \\ \hline
\multicolumn{8}{c}{\textit{pass@1}} \\ \hline
Qwen2.5-Math-7B + GRPO (Baseline) & 30.0 & 10.0 & 62.5 & 80.7 & 39.4 & 43.1 & 44.3 \\
+ LP-Weighting & 36.7 & \textbf{16.7} & 64.1 & 81.0 & 39.6 & 42.9 & 46.8 \\
+ LP-Weighting + PG-Sampling ($\beta_{min}$=0.1, $\beta_{max}$=0.5) & 36.7 & 13.3 & 70.0 & 81.9 & 40.0 & 44.4 & 47.7 \\
+ LP-Weighting + PG-Sampling ($\beta_{min}$=0.2, $\beta_{max}$=0.65) & 36.7 & 13.3 & 65.8 & 83.4 & 41.9 & 43.7 & 47.5 \\
+ LP-Weighting + PG-Sampling ($\beta_{min}$=0.3, $\beta_{max}$=0.8) & \textbf{40.0} & \textbf{16.7} & 69.2 & 82.0 & 41.5 & 43.5 & 48.8 \\ \hline
\multicolumn{8}{c}{\textit{pass@3}} \\ \hline
Qwen2.5-Math-7B + GRPO (Baseline) & 30.0 & 10.0 & 67.5 & 82.0 & 40.6 & 46.2 & 46.1 \\
+ LP-Weighting & 36.7 & \textbf{16.7} & 70.0 & 84.0 & 43.5 & 46.8 & 49.6 \\
+ LP-Weighting + PG-Sampling ($\beta_{min}$=0.1, $\beta_{max}$=0.5) & 36.7 & 13.3 & \textbf{75.0} & 83.2 & 42.3 & \textbf{47.1} & 49.6 \\
+ LP-Weighting + PG-Sampling ($\beta_{min}$=0.2, $\beta_{max}$=0.65) & 36.7 & \textbf{16.7} & 72.5 & \textbf{84.4} & 41.9 & 46.0 & 49.7 \\
+ LP-Weighting + PG-Sampling ($\beta_{min}$=0.3, $\beta_{max}$=0.8) & \textbf{40.0} & \textbf{16.7} & 72.5 & 83.8 & \textbf{46.0} & 46.6 & \textbf{50.9} \\ \hline
\end{tabular}
}
\caption{Comparison of model performance across six mathematical reasoning benchmarks. Results are reported using \textit{pass@1} and \textit{pass@3} metrics for different variants of PG-Sampling strategies.}
\label{tab:pg-sampling}
\end{table*}

\begin{figure*}[ht]
  \centering
  % ==== 四张子图 =============================================================
  \begin{subfigure}[b]{0.245\textwidth}
    \includegraphics[width=\linewidth]{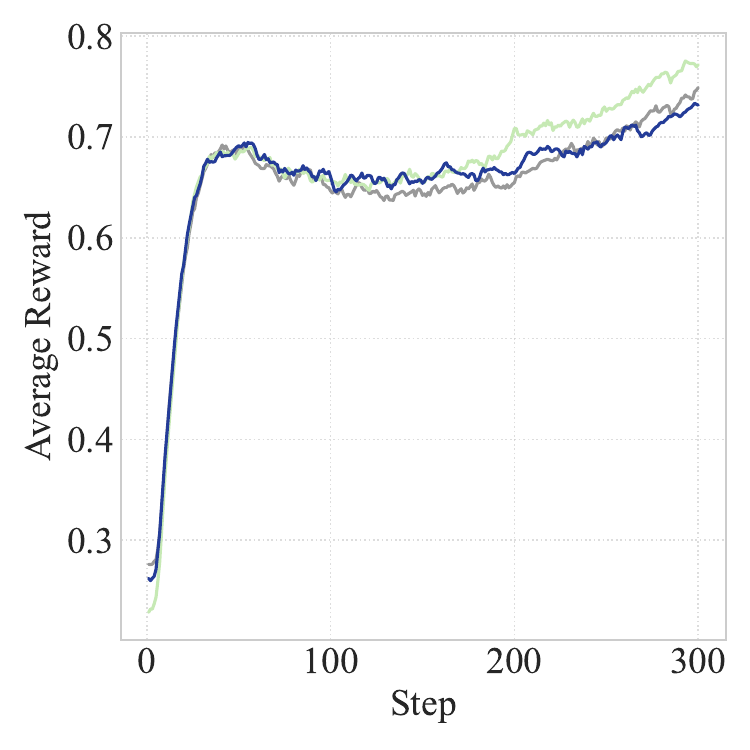}
    \caption{Average Reward}
    \label{fig:lp_sub1}
  \end{subfigure}
  \hfill
  \begin{subfigure}[b]{0.245\textwidth}
    \includegraphics[width=\linewidth]{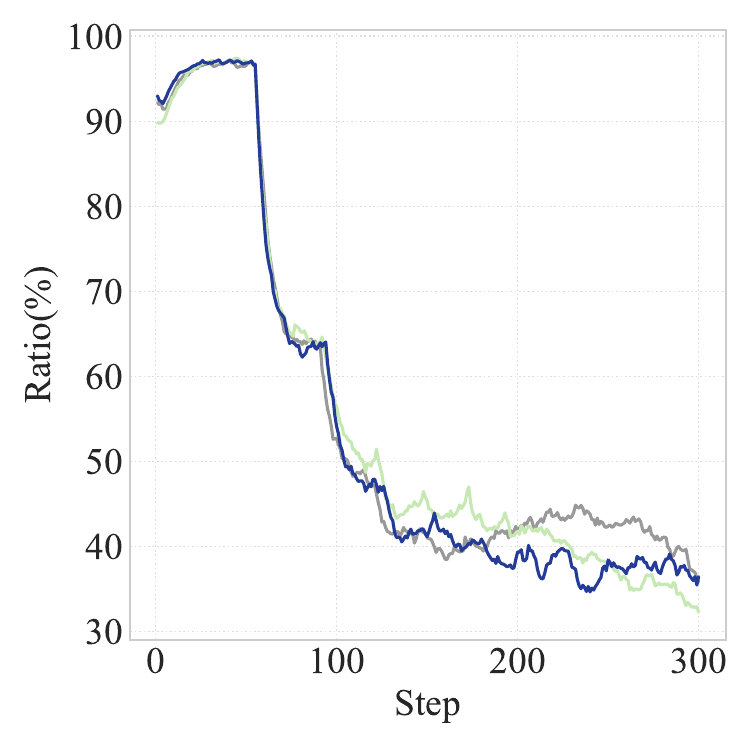}
    \caption{Improving Sample(\%)}
    \label{fig:lp_sub2}
  \end{subfigure}
  \hfill
  \begin{subfigure}[b]{0.245\textwidth}
    \includegraphics[width=\linewidth]{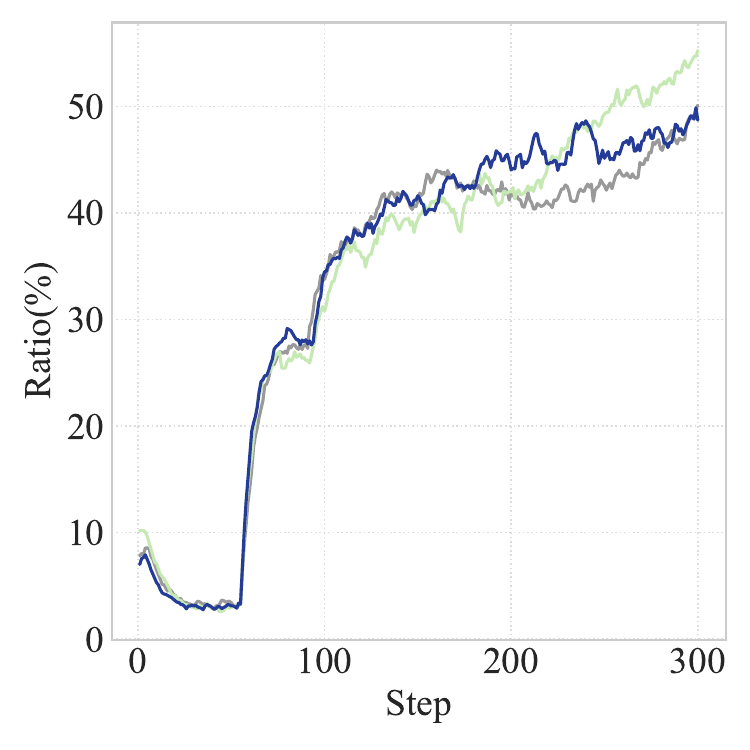}
    \caption{Plateauing Sample(\%)}
    \label{fig:lp_sub3}
  \end{subfigure}
  \hfill
  \begin{subfigure}[b]{0.245\textwidth}
    \includegraphics[width=\linewidth]{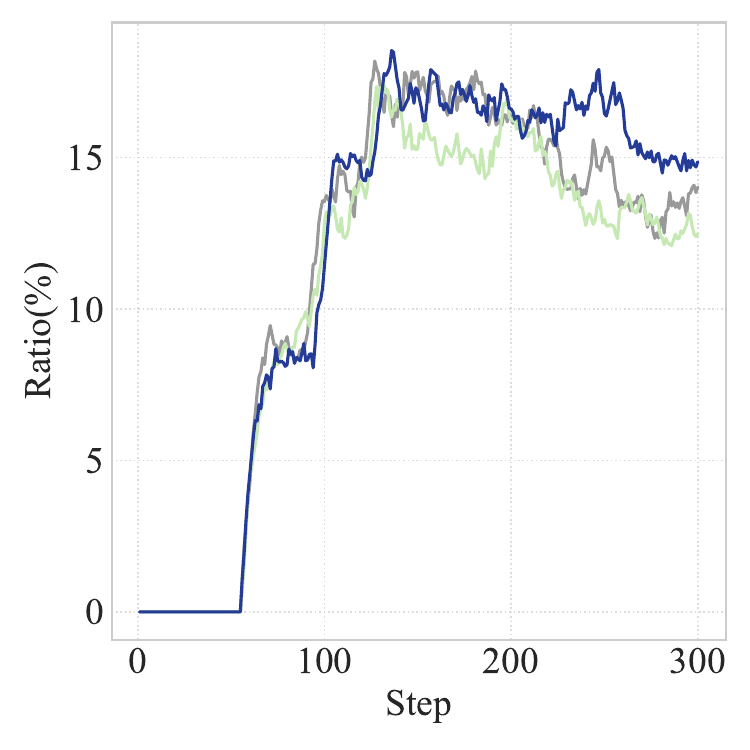}
    \caption{Degrading Sample(\%)}
    \label{fig:lp_sub4}
  \end{subfigure}

  % ======================= Legend ===========================================
  \vspace{0.2em} % 与子图稍作间隔
  \begin{minipage}{0.95\linewidth}
    \centering
    \legitem{gray!70}{\small GRPO}%
    \hspace{1em}% 调整不同图例项之间的水平间距
    \legitem{lpgreen}{\small GRPO + LP-Weighting}%
    \hspace{1em}
    \legitem{pgdeepblue}{\small GRPO + LP-Weighting (reverse)}%
  \end{minipage}

  \caption{Training dynamics under GRPO: (a) average reward, and (b–d) proportions of improving, plateauing, and degrading samples for the baseline GRPO, GRPO + LP-Weighting, and GRPO with reversed LP-Weighting.}
  \label{fig:lp_ablation}
\end{figure*}

\paragraph{LP-Weighting Can Boost Training}
Fig.~\ref{fig:lp_ablation}(a) shows that models trained with LP-Weighting achieve higher average rewards over time than both the GRPO baseline and the reversed-weighting variant, indicating more effective learning. Fig.~\ref{fig:lp_ablation}(b) shows that under LP-Weighting the fraction of improving samples steadily decreases, suggesting the model converges as more samples are learned. Conversely, Fig.~\ref{fig:lp_ablation}(c) shows that the proportion of plateauing samples (those with stable performance) rises with LP-Weighting, reflecting that many examples have reached a learning plateau. In Fig.~\ref{fig:lp_ablation}(d), the proportion of degrading samples (those losing performance) falls under LP-Weighting, implying increased robustness and fewer regressions. Together, these results indicate that LP-Weighting helps the model retain high-pass-rate samples and accelerates training by prioritizing samples with active learning signals.

\paragraph{Diversity} We examine how PG-Sampling affects the diversity of generated solutions using \textit{pass@$k$} metrics. Table~\ref{tab:pg-sampling} reports both \textit{pass@1} (\textit{avg@3}) and \textit{pass@3}. All methods achieve higher \textit{pass@3} than \textit{pass@1}, as expected, and PG-Sampling does not collapse this gap. In fact, LP-Weighting alone already widens the \textit{pass@3} margin over the baseline, and adding prefix-guided sampling maintains or even slightly enlarges it. For example, the $(0.3,0.8)$ PG-Sampling setting boosts \textit{pass@3} in line with \textit{pass@1}, indicating that multiple distinct correct answers remain available. In short, PG-Sampling preserves rich solution diversity: its accuracy gains come with sustained improvements in \textit{pass@3}, showing that the model continues to generate a variety of valid solutions for each problem.

\paragraph{Impact of $\beta$} We investigate how the truncation ratio range $(\beta_{\min},\beta_{\max})$ in PG-Sampling affects accuracy. As shown in Table~\ref{tab:pg-sampling}, all PG-Sampling variants with LP-Weighting outperform both the RLVR baseline and LP-Weighting alone. Expanding the prefix range generally improves results, with the widest range $(0.3,0.8)$ achieving the best performance. Narrower ranges still help but less so. This indicates that longer prefixes better guide the model, though gains taper off at extremes. In practice, a moderate-to-large $\beta$ range provides the best balance between effectiveness and exploration.

\paragraph{Impact of $\kappa$} 
We sweep the slope multiplier $\kappa\!\in\!\{4,8,12\}$, analytically fixing the bias at $b=0.5$ so that the expected weight $\mathbb{E}[w_i]$ in Eq.\ref{eq:dynamic_weight} is centred around~1—preventing either easy or hard samples from being systematically over- or under-emphasised.  
As Table~\ref{tab:kappa} shows, every LP-Weighting variant surpasses the RLVR baseline, and the macro average changes by at most 0.7 pp, demonstrating that LP-Weighting is largely insensitive to this hyper-parameter.  

\begin{table*}[t]
  \centering
  \resizebox{\textwidth}{!}{
\begin{tabular}{lccccccc}
\hline
Method / $\kappa$             & AIME24 & AIME25 & AMC23 & MATH-500 & Minerva & OlympiadBench & Avg           \\ \hline
Qwen2.5-Math-7B + GRPO (Baseline)   & 30.0   & 10.0   & 62.5  & 80.7     & 39.4    & 43.1          & 44.3          \\ \hline
+\,PG-Sampling + LP-Weighting ($\kappa$=4)  & \textbf{43.3}   & \textbf{16.7}   & 65.0  & 82.2     & \textbf{42.2}    & \textbf{44.6}          & \textbf{49.0}          \\
+\,PG-Sampling + LP-Weighting ($\kappa$=8)  & 40.0   & \textbf{16.7}   & \textbf{69.2}  & 82.0     & 41.5    & 43.5          & 48.8 \\
+\,PG-Sampling + LP-Weighting ($\kappa$=12) & 40.0   & 13.3   & 68.3  & \textbf{82.8}     & 41.5    & 44.0          & 48.3          \\ \hline
\end{tabular}
}
\caption{Effect of the LP-Weighting scaling factor $\kappa$ on pass@1 metric.}
\label{tab:kappa}
\end{table*}

\paragraph{LPPO under Diverse Scenarios.}
\label{sec:lppo_diverse}
To verify that our LPPO generalises beyond the original
Qwen-2.5-Math-7B setup, we replicate RLVR training in three
orthogonal settings without retuning any hyper-parameters:
(i) a larger backbone with Qwen-2.5-14B,
(ii) a different backbone family using Llama-3.2-3B-Instruct,
and (iii) an alternative policy-gradient learner (REINFORCE++).
Table~\ref{tab:w3_overview} summarises the results, while per-benchmark
breakdowns are deferred to Appendix~\ref{app:appendix_diverse}.
Across all variants, LPPO consistently yields \mbox{+2–4 pp} absolute
improvements in \textit{pass@1}, demonstrating robustness to model scale,
backbone architecture, and RL optimiser choice.

\begin{table}[t]
  \centering
  \resizebox{0.5\textwidth}{!}{
  \begin{tabular}{lccc}
    \hline
    Setting & Baseline & +LPPO & $\Delta$ (pp) \\\hline
    Qwen-2.5-14B \,+\, GRPO            & 42.7 & 45.0 & +2.3 \\
    Llama-3.2-3B-Instruct \,+\, GRPO   & 25.3 & 27.9 & +2.6 \\
    Qwen-2.5-Math-7B \,+\, REINFORCE++      & 44.6 & 48.7 & +4.1 \\\hline
  \end{tabular}}
  \caption{Robustness of LPPO across scale, architecture, and learner.
  Without extra tuning, LPPO consistently boosts \textit{pass@1} over larger
  parameter budgets, alternative backbones, and REINFORCE-style learners.}
  \label{tab:w3_overview}
\end{table}

%从两个角度论证：\ref{lp_ratio} a展示了训练过程中LP-Weighting，baseline，LP-Weighting—reverse（使用反向权重）的平均reward，可以观察到LP-Weighting在训练中拥有最高的reward。同时我们观察了三者取得积极进展improved-sample、停滞不前sample和衰退degraded-sample的比例（我们以训练中sample的pass rate进行衡量），如图\ref{lp_ratio} a、b和c所示。随着训练进行，LP-Weighting能够在更多sample上取得正向进展——即令通过率提高的样本的比例减小，而通过率维持不变（pass rate变换在正负5%以内），以及通过率降低的样本比例，其比例下降的样本在稳定降低，同时稳定样本比例持续上升。综上所属，可以推断出其将多数训练样本维持在了较高的pass rate水平，也就是在训练中更快取得进展，也和该方法的初衷一致——通过动态加权正在改进的样本权重加速训练。

% \paragraph{Online Data Curation in Practice}
% Online data curation improves training efficiency by \emph{continuously} filtering out solved and hard samples. Removing these cases can make computation focus on the remaining, informative examples.  
% Because the same wall-clock budget is now spent on fewer samples, the model revisits each useful sample more often, and this extra exposure also gives \textit{LP-Weighting} enough passes to take full effect.
% Fig.~\ref{fig:online_data_curation} highlights the impact of this process.  
% After about 60 steps, the active set has dropped to roughly 85\% of its initial size, and by step~300 it falls below 70\%, while the baseline remains at 100\%.  
% This \(\sim\!1.3\text{–}1.5\times\) boost in effective epoch count can make faster convergence and cooperate with LP-Weighting.

\begin{figure}[ht]
    \centering
    \includegraphics[width=0.38\textwidth]{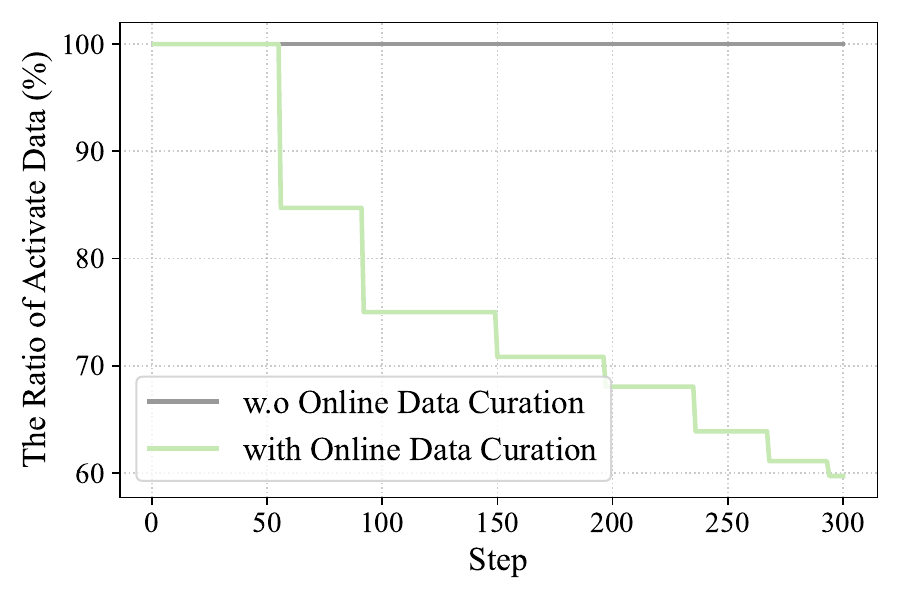} % 替换路径
    \caption{The Ratio of Activate Data during Training.}
    \label{fig:online_data_curation}
\end{figure}

\paragraph{Efficiency Gains via Online Data Curation}
Online data curation dynamically excludes both solved and overly difficult samples, allowing computation to focus on informative ones.
Each useful sample is revisited more frequently within the same wall-clock time, amplifying the effect of LP-Weighting.
As illustrated in Fig.~\ref{fig:online_data_curation}, this process narrows the active set over time compared to the baseline, leading to faster convergence and improved synergy with LP-Weighting.

% \paragraph{Balance between SFT and RL}
%此处可给出测试集中具有差异化的题目，并且给出度量prefix训练后样本 和原始solution相似度增加的结论。
% \paragraph{论证prefix方法能够改变输出分布/prefix方法使得模型具备更好的探索能力？通过KL进行}
% % 目前，RL是否能够提升模型的推理上界，还并未有一个共同的学界共识。当前基于RLVR的推理能力训练，是通过对同一个样本进行多次rollout后，以强化正确路径出现概率作为policy的更新方向，即（可引入公式）。这种更新可被近似于令pass@k逼近\textit{pass@1}。尽管能够提高模型的推理性能，但其探索空间仍旧被限制在base model之中。
%论证在同一个数据集（with solution）下，相同训练轮次下，SFT的泛化性弱于RL+prefix

%针对LP-Weighting策略的分析
% \paragraph{可能的和LPPO相关的实验,如反向的LPPO——让取得进步的样本权重降低，停滞不前的升高，是否会限制性能？或者让性能衰退？}
%\paragraph{LP-Weighting能够在当前有限步数下(仅300step)得到足够的训练吗}

\section{Conclusion}
We present \textbf{LPPO}, a sample-centric framework for reinforcement learning with verifiable rewards.  
LPPO unites \emph{learning-progress weighting}, which reallocates credit to examples that still drive improvement, with \emph{prefix-guided sampling}, which appends concise prefix hints only to problems that the current policy cannot yet solve.  
By concentrating computation on the most informative instances and injecting guidance on demand, LPPO shortens training time and significantly outperforms all baselines on Qwen-2.5-Math-7B.  
Furthermore, it delivers consistent 2–4\,pp gains in pass@1 accuracy across larger backbones, diverse architectures, and alternative policy-gradient learners.  
These results demonstrate that fine-grained, sample-level control offers a practical and widely applicable route to stronger mathematical reasoning LLMs.

% With a sample-centric perspective rooted in progressive optimization, we introduced PG-Sampling and LP-Weighting for mathematical reasoning. Our methods prioritize valuable training examples and leverage expert solutions, leading to faster convergence and higher accuracy. Our results demonstrate that sample-centric progressive optimization offers an effective RLVR framework for enhancing mathematical reasoning.

% In this work, we introduce two sample-centric RLVR techniques for mathematical reasoning. PG-Sampling guides exploration by appending expert solution prefixes, while LP-Weighting dynamically emphasizes samples showing active learning progress. Experiments demonstrate consistent improvement over strong RLVR baselines, with the combined methods achieving notably faster convergence and higher accuracy across benchmarks. Our approach effectively leverages expert demonstrations and adaptive weighting, efficiently targeting valuable examples for enhanced reasoning performance.

\section*{Limitations}
While our LPPO demonstrates clear advantages on mathematical reasoning benchmarks, several limitations remain. First, our experiments are conducted primarily on mathematical reasoning tasks with relatively small, high-quality expert-annotated datasets; the generalizability of these methods to broader reasoning domains or more diverse tasks requires further investigation. Second, PG-Sampling relies on the availability of expert solutions for challenging problems, which may not always be feasible for other tasks. Third, although our methods improve training efficiency and convergence speed, they introduce additional complexity in data preparation and tracking per-sample statistics. Lastly, the overall improvements are bounded by the underlying model capacity and the difficulty of the evaluation benchmarks, and may not directly translate to real-world applications. We leave these directions for future work.

\bibliography{custom}

\begin{thebibliography}{60}
\providecommand{\natexlab}[1]{#1}

\bibitem[{Aggarwal and Welleck(2025)}]{1.18_reward_l1}
Pranjal Aggarwal and Sean Welleck. 2025.
\newblock \href {https://doi.org/10.48550/ARXIV.2503.04697} {{L1:} controlling how long {A} reasoning model thinks with reinforcement learning}.
\newblock \emph{CoRR}, abs/2503.04697.

\bibitem[{Ahmadian et~al.(2024)Ahmadian, Cremer, Gall{\'{e}}, Fadaee, Kreutzer, Pietquin, {\"{U}}st{\"{u}}n, and Hooker}]{1.8_REINFORCE}
Arash Ahmadian, Chris Cremer, Matthias Gall{\'{e}}, Marzieh Fadaee, Julia Kreutzer, Olivier Pietquin, Ahmet {\"{U}}st{\"{u}}n, and Sara Hooker. 2024.
\newblock \href {https://doi.org/10.18653/V1/2024.ACL-LONG.662} {Back to basics: Revisiting reinforce-style optimization for learning from human feedback in llms}.
\newblock In \emph{Proceedings of the 62nd Annual Meeting of the Association for Computational Linguistics (Volume 1: Long Papers), {ACL} 2024, Bangkok, Thailand, August 11-16, 2024}, pages 12248--12267. Association for Computational Linguistics.

\bibitem[{Albalak et~al.(2025)Albalak, Phung, Lile, Rafailov, Gandhi, Castricato, Singh, Blagden, Xiang, Mahan, and Haber}]{1.14_data_bigmath}
Alon Albalak, Duy Phung, Nathan Lile, Rafael Rafailov, Kanishk Gandhi, Louis Castricato, Anikait Singh, Chase Blagden, Violet Xiang, Dakota Mahan, and Nick Haber. 2025.
\newblock \href {https://arxiv.org/abs/2502.17387} {Big-math: A large-scale, high-quality math dataset for reinforcement learning in language models}.
\newblock \emph{Preprint}, arXiv:2502.17387.

\bibitem[{Bengio et~al.(2009)Bengio, Louradour, Collobert, and Weston}]{3_1.7_curriculum}
Yoshua Bengio, J{\'{e}}r{\^{o}}me Louradour, Ronan Collobert, and Jason Weston. 2009.
\newblock \href {https://doi.org/10.1145/1553374.1553380} {Curriculum learning}.
\newblock In \emph{Proceedings of the 26th Annual International Conference on Machine Learning, {ICML} 2009, Montreal, Quebec, Canada, June 14-18, 2009}, volume 382 of \emph{{ACM} International Conference Proceeding Series}, pages 41--48. {ACM}.

\bibitem[{Chen et~al.(2024{\natexlab{a}})Chen, Liao, Li, and Fan}]{3_1.1_supermario}
Guoxin Chen, Minpeng Liao, Chengxi Li, and Kai Fan. 2024{\natexlab{a}}.
\newblock \href {http://papers.nips.cc/paper\_files/paper/2024/hash/30dfe47a3ccbee68cffa0c19ccb1bc00-Abstract-Conference.html} {Alphamath almost zero: Process supervision without process}.
\newblock In \emph{Advances in Neural Information Processing Systems 38: Annual Conference on Neural Information Processing Systems 2024, NeurIPS 2024, Vancouver, BC, Canada, December 10 - 15, 2024}.

\bibitem[{Chen et~al.(2024{\natexlab{b}})Chen, Li, Yan, Wang, Gunaratna, Yadav, Tang, Srinivasan, Zhou, Huang, and Jin}]{2.7_data_AlpaGasus}
Lichang Chen, Shiyang Li, Jun Yan, Hai Wang, Kalpa Gunaratna, Vikas Yadav, Zheng Tang, Vijay Srinivasan, Tianyi Zhou, Heng Huang, and Hongxia Jin. 2024{\natexlab{b}}.
\newblock \href {https://openreview.net/forum?id=FdVXgSJhvz} {Alpagasus: Training a better alpaca with fewer data}.
\newblock In \emph{The Twelfth International Conference on Learning Representations, {ICLR} 2024, Vienna, Austria, May 7-11, 2024}. OpenReview.net.

\bibitem[{Chu et~al.(2025)Chu, Huang, Zhang, Wei, and Wang}]{4_6_GPG}
Xiangxiang Chu, Hailang Huang, Xiao Zhang, Fei Wei, and Yong Wang. 2025.
\newblock \href {https://arxiv.org/abs/2504.02546} {Gpg: A simple and strong reinforcement learning baseline for model reasoning}.
\newblock \emph{Preprint}, arXiv:2504.02546.

\bibitem[{Cui et~al.(2025)Cui, Yuan, Wang, Wang, Li, He, Fan, Yu, Xu, Chen, Yuan, Chen, Zhang, Lv, Wang, Yao, Han, Peng, Cheng, Liu, Sun, Zhou, and Ding}]{4_10_Eurus_prime}
Ganqu Cui, Lifan Yuan, Zefan Wang, Hanbin Wang, Wendi Li, Bingxiang He, Yuchen Fan, Tianyu Yu, Qixin Xu, Weize Chen, Jiarui Yuan, Huayu Chen, Kaiyan Zhang, Xingtai Lv, Shuo Wang, Yuan Yao, Xu~Han, Hao Peng, Yu~Cheng, and 4 others. 2025.
\newblock \href {https://doi.org/10.48550/ARXIV.2502.01456} {Process reinforcement through implicit rewards}.
\newblock \emph{CoRR}, abs/2502.01456.

\bibitem[{Das et~al.(2024)Das, Chakraborty, Pacchiano, and Chowdhury}]{2.10_2_data}
Nirjhar Das, Souradip Chakraborty, Aldo Pacchiano, and Sayak~Ray Chowdhury. 2024.
\newblock \href {https://arxiv.org/abs/2402.10500} {Active preference optimization for sample efficient rlhf}.
\newblock \emph{Preprint}, arXiv:2402.10500.

\bibitem[{DeepSeek{-}AI et~al.(2025)DeepSeek{-}AI, Guo, Yang, Zhang, Song, Zhang, Xu, Zhu, Ma, Wang, and Xiao~Bi}]{1.1_deepseek_r1}
DeepSeek{-}AI, Daya Guo, Dejian Yang, Haowei Zhang, Junxiao Song, Ruoyu Zhang, Runxin Xu, Qihao Zhu, Shirong Ma, Peiyi Wang, and et~al Xiao~Bi. 2025.
\newblock \href {https://doi.org/10.48550/ARXIV.2501.12948} {Deepseek-r1: Incentivizing reasoning capability in llms via reinforcement learning}.
\newblock \emph{CoRR}, abs/2501.12948.

\bibitem[{Fatemi et~al.(2025)Fatemi, Rafiee, Tang, and Talamadupula}]{2.14_4sample}
Mehdi Fatemi, Banafsheh Rafiee, Mingjie Tang, and Kartik Talamadupula. 2025.
\newblock \href {https://arxiv.org/abs/2504.05185} {Concise reasoning via reinforcement learning}.
\newblock \emph{Preprint}, arXiv:2504.05185.

\bibitem[{Gao et~al.(2024{\natexlab{a}})Gao, Xu, Ye, Liu, He, Fu, Mei, Wang, and Wu}]{1.3_RLVR}
Jiaxuan Gao, Shusheng Xu, Wenjie Ye, Weilin Liu, Chuyi He, Wei Fu, Zhiyu Mei, Guangju Wang, and Yi~Wu. 2024{\natexlab{a}}.
\newblock \href {https://doi.org/10.48550/ARXIV.2410.15115} {On designing effective {RL} reward at training time for {LLM} reasoning}.
\newblock \emph{CoRR}, abs/2410.15115.

\bibitem[{Gao et~al.(2024{\natexlab{b}})Gao, Xu, Ye, Liu, He, Fu, Mei, Wang, and Wu}]{2.1_rlvr}
Jiaxuan Gao, Shusheng Xu, Wenjie Ye, Weilin Liu, Chuyi He, Wei Fu, Zhiyu Mei, Guangju Wang, and Yi~Wu. 2024{\natexlab{b}}.
\newblock \href {https://doi.org/10.48550/ARXIV.2410.15115} {On designing effective {RL} reward at training time for {LLM} reasoning}.
\newblock \emph{CoRR}, abs/2410.15115.

\bibitem[{He et~al.(2024)He, Luo, Bai, Hu, Thai, Shen, Hu, Han, Huang, Zhang, Liu, Qi, Liu, and Sun}]{4_5_OlympiadBench}
Chaoqun He, Renjie Luo, Yuzhuo Bai, Shengding Hu, Zhen~Leng Thai, Junhao Shen, Jinyi Hu, Xu~Han, Yujie Huang, Yuxiang Zhang, Jie Liu, Lei Qi, Zhiyuan Liu, and Maosong Sun. 2024.
\newblock \href {https://doi.org/10.18653/V1/2024.ACL-LONG.211} {Olympiadbench: {A} challenging benchmark for promoting {AGI} with olympiad-level bilingual multimodal scientific problems}.
\newblock In \emph{Proceedings of the 62nd Annual Meeting of the Association for Computational Linguistics (Volume 1: Long Papers), {ACL} 2024, Bangkok, Thailand, August 11-16, 2024}, pages 3828--3850. Association for Computational Linguistics.

\bibitem[{Hendrycks et~al.(2021)Hendrycks, Burns, Kadavath, Arora, Basart, Tang, Song, and Steinhardt}]{4_3_math_dataset}
Dan Hendrycks, Collin Burns, Saurav Kadavath, Akul Arora, Steven Basart, Eric Tang, Dawn Song, and Jacob Steinhardt. 2021.
\newblock \href {https://datasets-benchmarks-proceedings.neurips.cc/paper/2021/hash/be83ab3ecd0db773eb2dc1b0a17836a1-Abstract-round2.html} {Measuring mathematical problem solving with the {MATH} dataset}.
\newblock In \emph{Proceedings of the Neural Information Processing Systems Track on Datasets and Benchmarks 1, NeurIPS Datasets and Benchmarks 2021, December 2021, virtual}.

\bibitem[{Hochlehnert et~al.(2025)Hochlehnert, Bhatnagar, Udandarao, Albanie, Prabhu, and Bethge}]{4_7_sober_repro_eval}
Andreas Hochlehnert, Hardik Bhatnagar, Vishaal Udandarao, Samuel Albanie, Ameya Prabhu, and Matthias Bethge. 2025.
\newblock \href {https://arxiv.org/abs/2504.07086} {A sober look at progress in language model reasoning: Pitfalls and paths to reproducibility}.
\newblock \emph{Preprint}, arXiv:2504.07086.

\bibitem[{Hoffmann et~al.(2022)Hoffmann, Borgeaud, Mensch, Buchatskaya, Cai, Rutherford, de~Las~Casas, Hendricks, Welbl, Clark, Hennigan, Noland, Millican, van~den Driessche, Damoc, Guy, Osindero, Simonyan, Elsen, Rae, Vinyals, and Sifre}]{3_1.6_scaling_law}
Jordan Hoffmann, Sebastian Borgeaud, Arthur Mensch, Elena Buchatskaya, Trevor Cai, Eliza Rutherford, Diego de~Las~Casas, Lisa~Anne Hendricks, Johannes Welbl, Aidan Clark, Tom Hennigan, Eric Noland, Katie Millican, George van~den Driessche, Bogdan Damoc, Aurelia Guy, Simon Osindero, Karen Simonyan, Erich Elsen, and 3 others. 2022.
\newblock \href {https://doi.org/10.48550/ARXIV.2203.15556} {Training compute-optimal large language models}.
\newblock \emph{CoRR}, abs/2203.15556.

\bibitem[{Hu(2025)}]{1.12_REINFORCE++}
Jian Hu. 2025.
\newblock \href {https://doi.org/10.48550/ARXIV.2501.03262} {{REINFORCE++:} {A} simple and efficient approach for aligning large language models}.
\newblock \emph{CoRR}, abs/2501.03262.

\bibitem[{Hu et~al.(2025)Hu, Zhang, Han, Jiang, Zhang, and Shum}]{1.13_data_orz}
Jingcheng Hu, Yinmin Zhang, Qi~Han, Daxin Jiang, Xiangyu Zhang, and Heung-Yeung Shum. 2025.
\newblock \href {https://arxiv.org/abs/2503.24290} {Open-reasoner-zero: An open source approach to scaling up reinforcement learning on the base model}.
\newblock \emph{Preprint}, arXiv:2503.24290.

\bibitem[{Huang et~al.(2024)Huang, Zou, Li, Liu, Zheng, Chern, Xia, Qin, Yuan, and Liu}]{4_2_o1_journey}
Zhen Huang, Haoyang Zou, Xuefeng Li, Yixiu Liu, Yuxiang Zheng, Ethan Chern, Shijie Xia, Yiwei Qin, Weizhe Yuan, and Pengfei Liu. 2024.
\newblock \href {https://doi.org/10.48550/ARXIV.2411.16489} {{O1} replication journey - part 2: Surpassing o1-preview through simple distillation, big progress or bitter lesson?}
\newblock \emph{CoRR}, abs/2411.16489.

\bibitem[{Hunter(1986)}]{3_1.11_ema}
J~Stuart Hunter. 1986.
\newblock The exponentially weighted moving average.
\newblock \emph{Journal of quality technology}, 18(4):203--210.

\bibitem[{Ivison et~al.(2023)Ivison, Smith, Hajishirzi, and Dasigi}]{2.8_data}
Hamish Ivison, Noah~A. Smith, Hannaneh Hajishirzi, and Pradeep Dasigi. 2023.
\newblock \href {https://doi.org/10.18653/V1/2023.FINDINGS-ACL.576} {Data-efficient finetuning using cross-task nearest neighbors}.
\newblock In \emph{Findings of the Association for Computational Linguistics: {ACL} 2023, Toronto, Canada, July 9-14, 2023}, pages 9036--9061. Association for Computational Linguistics.

\bibitem[{Ivison et~al.(2025)Ivison, Zhang, Brahman, Koh, and Dasigi}]{2.6_data}
Hamish Ivison, Muru Zhang, Faeze Brahman, Pang~Wei Koh, and Pradeep Dasigi. 2025.
\newblock \href {https://doi.org/10.48550/ARXIV.2503.01807} {Large-scale data selection for instruction tuning}.
\newblock \emph{CoRR}, abs/2503.01807.

\bibitem[{Kaplan et~al.(2020)Kaplan, McCandlish, Henighan, Brown, Chess, Child, Gray, Radford, Wu, and Amodei}]{3_1.5_scaling_law}
Jared Kaplan, Sam McCandlish, Tom Henighan, Tom~B. Brown, Benjamin Chess, Rewon Child, Scott Gray, Alec Radford, Jeffrey Wu, and Dario Amodei. 2020.
\newblock \href {https://arxiv.org/abs/2001.08361} {Scaling laws for neural language models}.
\newblock \emph{CoRR}, abs/2001.08361.

\bibitem[{Kazemnejad et~al.(2024)Kazemnejad, Aghajohari, Portelance, Sordoni, Reddy, Courville, and Roux}]{2.5_2_vineppo}
Amirhossein Kazemnejad, Milad Aghajohari, Eva Portelance, Alessandro Sordoni, Siva Reddy, Aaron~C. Courville, and Nicolas~Le Roux. 2024.
\newblock \href {https://doi.org/10.48550/ARXIV.2410.01679} {Vineppo: Unlocking {RL} potential for {LLM} reasoning through refined credit assignment}.
\newblock \emph{CoRR}, abs/2410.01679.

\bibitem[{Kimi et~al.(2025)Kimi, Du, Gao, Xing, Jiang, Chen, Li, Xiao, Du, Liao, Tang, Wang, Zhang, Yuan, Lu, Tang, Sung, Wei, Lai, Guo, Zhu, Ding, Hu, Yang, Zhang, Yao, Zhao, Lu, Li, Yu, Gao, Zheng, Yuan, Chen, Guo, Su, Wang, Zhao, Zhang, Liu, Yan, Wu, Shi, Ye, Yu, Dong, Zhang, Ma, Pan, Gong, Liu, Ma, Wei, Cao, Huang, Jiang, Gao, Xiong, He, Huang, Wu, He, Wei, Jia, Wu, Xu, Zu, Zhou, Pan, Charles, Li, Hu, Liu, Chen, Wang, Liu, Qin, Liu, Yang, Bao, Du, Wu, Wang, Zhou, Wang, Li, Zhu, Zhang, Wang, Yang, Huang, Huang, Xu, and Yang}]{1.2_kimi_k1.5}
Kimi, Angang Du, Bofei Gao, Bowei Xing, Changjiu Jiang, Cheng Chen, Cheng Li, Chenjun Xiao, Chenzhuang Du, Chonghua Liao, Chuning Tang, Congcong Wang, Dehao Zhang, Enming Yuan, Enzhe Lu, Fengxiang Tang, Flood Sung, Guangda Wei, Guokun Lai, and 75 others. 2025.
\newblock \href {https://doi.org/10.48550/ARXIV.2501.12599} {Kimi k1.5: Scaling reinforcement learning with llms}.
\newblock \emph{CoRR}, abs/2501.12599.

\bibitem[{Kwon et~al.(2023)Kwon, Li, Zhuang, Sheng, Zheng, Yu, Gonzalez, Zhang, and Stoica}]{4_9_vllm}
Woosuk Kwon, Zhuohan Li, Siyuan Zhuang, Ying Sheng, Lianmin Zheng, Cody~Hao Yu, Joseph Gonzalez, Hao Zhang, and Ion Stoica. 2023.
\newblock \href {https://doi.org/10.1145/3600006.3613165} {Efficient memory management for large language model serving with pagedattention}.
\newblock In \emph{Proceedings of the 29th Symposium on Operating Systems Principles, {SOSP} 2023, Koblenz, Germany, October 23-26, 2023}, pages 611--626. {ACM}.

\bibitem[{Lewkowycz et~al.(2022)Lewkowycz, Andreassen, Dohan, Dyer, Michalewski, Ramasesh, Slone, Anil, Schlag, Gutman{-}Solo, Wu, Neyshabur, Gur{-}Ari, and Misra}]{4_4_Minerva_dataset}
Aitor Lewkowycz, Anders Andreassen, David Dohan, Ethan Dyer, Henryk Michalewski, Vinay~V. Ramasesh, Ambrose Slone, Cem Anil, Imanol Schlag, Theo Gutman{-}Solo, Yuhuai Wu, Behnam Neyshabur, Guy Gur{-}Ari, and Vedant Misra. 2022.
\newblock \href {http://papers.nips.cc/paper\_files/paper/2022/hash/18abbeef8cfe9203fdf9053c9c4fe191-Abstract-Conference.html} {Solving quantitative reasoning problems with language models}.
\newblock In \emph{Advances in Neural Information Processing Systems 35: Annual Conference on Neural Information Processing Systems 2022, NeurIPS 2022, New Orleans, LA, USA, November 28 - December 9, 2022}.

\bibitem[{Li et~al.(2024)Li, Beeching, Tunstall, Lipkin, Soletskyi, Huang, Rasul, Yu, Jiang, Shen et~al.}]{4_11_Numinamath}
Jia Li, Edward Beeching, Lewis Tunstall, Ben Lipkin, Roman Soletskyi, Shengyi Huang, Kashif Rasul, Longhui Yu, Albert~Q Jiang, Ziju Shen, and 1 others. 2024.
\newblock \href {http://faculty.bicmr.pku.edu.cn/~dongbin/Publications/numina_dataset.pdf} {Numinamath: The largest public dataset in ai4maths with 860k pairs of competition math problems and solutions}.
\newblock \emph{Hugging Face repository}, 13:9.

\bibitem[{Li et~al.(2025)Li, Zou, and Liu}]{1.15_data_limr}
Xuefeng Li, Haoyang Zou, and Pengfei Liu. 2025.
\newblock \href {https://doi.org/10.48550/ARXIV.2502.11886} {{LIMR:} less is more for {RL} scaling}.
\newblock \emph{CoRR}, abs/2502.11886.

\bibitem[{Lightman et~al.(2024)Lightman, Kosaraju, Burda, Edwards, Baker, Lee, Leike, Schulman, Sutskever, and Cobbe}]{4_4_math500_dataset}
Hunter Lightman, Vineet Kosaraju, Yuri Burda, Harrison Edwards, Bowen Baker, Teddy Lee, Jan Leike, John Schulman, Ilya Sutskever, and Karl Cobbe. 2024.
\newblock \href {https://openreview.net/forum?id=v8L0pN6EOi} {Let's verify step by step}.
\newblock In \emph{The Twelfth International Conference on Learning Representations, {ICLR} 2024, Vienna, Austria, May 7-11, 2024}. OpenReview.net.

\bibitem[{Liu et~al.(2025)Liu, Chen, Li, Qi, Pang, Du, Lee, and Lin}]{1.9_UnderstandingR1-Zero-LikeTraining_drgrpo}
Zichen Liu, Changyu Chen, Wenjun Li, Penghui Qi, Tianyu Pang, Chao Du, Wee~Sun Lee, and Min Lin. 2025.
\newblock \href {https://doi.org/10.48550/ARXIV.2503.20783} {Understanding r1-zero-like training: {A} critical perspective}.
\newblock \emph{CoRR}, abs/2503.20783.

\bibitem[{Muldrew et~al.(2024)Muldrew, Hayes, Zhang, and Barber}]{2.10_data}
William Muldrew, Peter Hayes, Mingtian Zhang, and David Barber. 2024.
\newblock \href {https://openreview.net/forum?id=CTgEV6qgUy} {Active preference learning for large language models}.
\newblock In \emph{Forty-first International Conference on Machine Learning, {ICML} 2024, Vienna, Austria, July 21-27, 2024}. OpenReview.net.

\bibitem[{Narvekar et~al.(2020)Narvekar, Peng, Leonetti, Sinapov, Taylor, and Stone}]{3_1.8_curriculum}
Sanmit Narvekar, Bei Peng, Matteo Leonetti, Jivko Sinapov, Matthew~E. Taylor, and Peter Stone. 2020.
\newblock \href {https://jmlr.org/papers/v21/20-212.html} {Curriculum learning for reinforcement learning domains: {A} framework and survey}.
\newblock \emph{J. Mach. Learn. Res.}, 21:181:1--181:50.

\bibitem[{Ouyang et~al.(2022)Ouyang, Wu, Jiang, Almeida, Wainwright, Mishkin, Zhang, Agarwal, Slama, Ray, Schulman, Hilton, Kelton, Miller, Simens, Askell, Welinder, Christiano, Leike, and Lowe}]{2.11_data}
Long Ouyang, Jeffrey Wu, Xu~Jiang, Diogo Almeida, Carroll~L. Wainwright, Pamela Mishkin, Chong Zhang, Sandhini Agarwal, Katarina Slama, Alex Ray, John Schulman, Jacob Hilton, Fraser Kelton, Luke Miller, Maddie Simens, Amanda Askell, Peter Welinder, Paul~F. Christiano, Jan Leike, and Ryan Lowe. 2022.
\newblock \href {http://papers.nips.cc/paper\_files/paper/2022/hash/b1efde53be364a73914f58805a001731-Abstract-Conference.html} {Training language models to follow instructions with human feedback}.
\newblock In \emph{Advances in Neural Information Processing Systems 35: Annual Conference on Neural Information Processing Systems 2022, NeurIPS 2022, New Orleans, LA, USA, November 28 - December 9, 2022}.

\bibitem[{Schulman et~al.(2017)Schulman, Wolski, Dhariwal, Radford, and Klimov}]{1.5_ppo}
John Schulman, Filip Wolski, Prafulla Dhariwal, Alec Radford, and Oleg Klimov. 2017.
\newblock \href {https://arxiv.org/abs/1707.06347} {Proximal policy optimization algorithms}.
\newblock \emph{CoRR}, abs/1707.06347.

\bibitem[{Shao et~al.(2024)Shao, Wang, Zhu, Xu, Song, Zhang, Li, Wu, and Guo}]{1.6_grpo}
Zhihong Shao, Peiyi Wang, Qihao Zhu, Runxin Xu, Junxiao Song, Mingchuan Zhang, Y.~K. Li, Y.~Wu, and Daya Guo. 2024.
\newblock \href {https://doi.org/10.48550/ARXIV.2402.03300} {Deepseekmath: Pushing the limits of mathematical reasoning in open language models}.
\newblock \emph{CoRR}, abs/2402.03300.

\bibitem[{Sheng et~al.(2025)Sheng, Zhang, Ye, Wu, Zhang, Zhang, Peng, Lin, and Wu}]{4_8_verl}
Guangming Sheng, Chi Zhang, Zilingfeng Ye, Xibin Wu, Wang Zhang, Ru~Zhang, Yanghua Peng, Haibin Lin, and Chuan Wu. 2025.
\newblock \href {https://doi.org/10.1145/3689031.3696075} {Hybridflow: {A} flexible and efficient {RLHF} framework}.
\newblock In \emph{Proceedings of the Twentieth European Conference on Computer Systems, EuroSys 2025, Rotterdam, The Netherlands, 30 March 2025 - 3 April 2025}, pages 1279--1297. {ACM}.

\bibitem[{Smirnov et~al.(2018)Smirnov, Melnikov, Oleinik, Ivanova, Kalinovskiy, and Luckyanets}]{3_1.10_hard_example_mining}
Evgeny Smirnov, Aleksandr Melnikov, Andrei Oleinik, Elizaveta Ivanova, Ilya Kalinovskiy, and Eugene Luckyanets. 2018.
\newblock \href {https://doi.org/10.1109/CVPRW.2018.00013} {Hard example mining with auxiliary embeddings}.
\newblock In \emph{2018 {IEEE} Conference on Computer Vision and Pattern Recognition Workshops, {CVPR} Workshops 2018, Salt Lake City, UT, USA, June 18-22, 2018}, pages 37--46. Computer Vision Foundation / {IEEE} Computer Society.

\bibitem[{Song et~al.(2025)Song, Zheng, Li, Yang, Luo, Pan, and Zhang}]{2.4_rlvr_cur_FastCuRL}
Mingyang Song, Mao Zheng, Zheng Li, Wenjie Yang, Xuan Luo, Yue Pan, and Feng Zhang. 2025.
\newblock \href {https://doi.org/10.48550/ARXIV.2503.17287} {Fastcurl: Curriculum reinforcement learning with progressive context extension for efficient training r1-like reasoning models}.
\newblock \emph{CoRR}, abs/2503.17287.

\bibitem[{Veeraboina(2023)}]{aime_1983_2023}
Hemish Veeraboina. 2023.
\newblock Aime problem set 1983--2024.
\newblock \url{https://www.kaggle.com/datasets/hemishveeraboina/aime-problem-set-1983-2024}.

\bibitem[{Wang et~al.(2020)Wang, Wei, Dong, Bao, Yang, and Zhou}]{4_12_sentence_transformer}
Wenhui Wang, Furu Wei, Li~Dong, Hangbo Bao, Nan Yang, and Ming Zhou. 2020.
\newblock \href {https://proceedings.neurips.cc/paper/2020/hash/3f5ee243547dee91fbd053c1c4a845aa-Abstract.html} {Minilm: Deep self-attention distillation for task-agnostic compression of pre-trained transformers}.
\newblock In \emph{Advances in Neural Information Processing Systems 33: Annual Conference on Neural Information Processing Systems 2020, NeurIPS 2020, December 6-12, 2020, virtual}.

\bibitem[{Wang et~al.(2025)Wang, Yang, Zeng, Ren, Liu, Peng, Cheng, He, Wang, Gao, Chen, Wang, Du, and Shen}]{2_3.5_one_sample}
Yiping Wang, Qing Yang, Zhiyuan Zeng, Liliang Ren, Lucas Liu, Baolin Peng, Hao Cheng, Xuehai He, Kuan Wang, Jianfeng Gao, Weizhu Chen, Shuohang Wang, Simon~Shaolei Du, and Yelong Shen. 2025.
\newblock \href {https://arxiv.org/abs/2504.20571} {Reinforcement learning for reasoning in large language models with one training example}.
\newblock \emph{Preprint}, arXiv:2504.20571.

\bibitem[{Wen et~al.(2025)Wen, Cai, Xiao, He, An, Duan, Du, Liu, Tang, Lv, Zou, Deng, Jia, and Zhang}]{2.3_rlvr_cur}
Liang Wen, Yunke Cai, Fenrui Xiao, Xin He, Qi~An, Zhenyu Duan, Yimin Du, Junchen Liu, Lifu Tang, Xiaowei Lv, Haosheng Zou, Yongchao Deng, Shousheng Jia, and Xiangzheng Zhang. 2025.
\newblock \href {https://doi.org/10.48550/ARXIV.2503.10460} {Light-r1: Curriculum sft, {DPO} and {RL} for long {COT} from scratch and beyond}.
\newblock \emph{CoRR}, abs/2503.10460.

\bibitem[{Xi et~al.(2024)Xi, Chen, Hong, Jin, Zheng, He, Ding, Liu, Guo, Wang, Guo, Shen, Fan, Zhou, Dou, Wang, Zhang, Sun, Gui, Zhang, and Huang}]{r3_add}
Zhiheng Xi, Wenxiang Chen, Boyang Hong, Senjie Jin, Rui Zheng, Wei He, Yiwen Ding, Shichun Liu, Xin Guo, Junzhe Wang, Honglin Guo, Wei Shen, Xiaoran Fan, Yuhao Zhou, Shihan Dou, Xiao Wang, Xinbo Zhang, Peng Sun, Tao Gui, and 2 others. 2024.
\newblock \href {https://openreview.net/forum?id=t82Y3fmRtk} {Training large language models for reasoning through reverse curriculum reinforcement learning}.
\newblock In \emph{Forty-first International Conference on Machine Learning, {ICML} 2024, Vienna, Austria, July 21-27, 2024}. OpenReview.net.

\bibitem[{Xia et~al.(2024)Xia, Malladi, Gururangan, Arora, and Chen}]{2.9_data}
Mengzhou Xia, Sadhika Malladi, Suchin Gururangan, Sanjeev Arora, and Danqi Chen. 2024.
\newblock \href {https://openreview.net/forum?id=PG5fV50maR} {{LESS:} selecting influential data for targeted instruction tuning}.
\newblock In \emph{Forty-first International Conference on Machine Learning, {ICML} 2024, Vienna, Austria, July 21-27, 2024}. OpenReview.net.

\bibitem[{Xie et~al.(2025)Xie, Gao, Ren, Luo, Hong, Dai, Zhou, Qiu, Wu, and Luo}]{1.17_data_logicRL}
Tian Xie, Zitian Gao, Qingnan Ren, Haoming Luo, Yuqian Hong, Bryan Dai, Joey Zhou, Kai Qiu, Zhirong Wu, and Chong Luo. 2025.
\newblock \href {https://doi.org/10.48550/ARXIV.2502.14768} {Logic-rl: Unleashing {LLM} reasoning with rule-based reinforcement learning}.
\newblock \emph{CoRR}, abs/2502.14768.

\bibitem[{Xu et~al.(2013)Xu, Sun, and Zhang}]{3_1.8_active_learning}
Yan Xu, Fuming Sun, and Xue Zhang. 2013.
\newblock \href {https://doi.org/10.1145/2499788.2499794} {Literature survey of active learning in multimedia annotation and retrieval}.
\newblock In \emph{International Conference on Internet Multimedia Computing and Service, {ICIMCS} '13, Huangshan, China - August 17 - 19, 2013}, pages 237--242. {ACM}.

\bibitem[{Yang et~al.(2024)Yang, Zhang, Hui, Gao, Yu, Li, Liu, Tu, Zhou, Lin, Lu, Xue, Lin, Liu, Ren, and Zhang}]{4_1_qwen_math}
An~Yang, Beichen Zhang, Binyuan Hui, Bofei Gao, Bowen Yu, Chengpeng Li, Dayiheng Liu, Jianhong Tu, Jingren Zhou, Junyang Lin, Keming Lu, Mingfeng Xue, Runji Lin, Tianyu Liu, Xingzhang Ren, and Zhenru Zhang. 2024.
\newblock \href {https://doi.org/10.48550/ARXIV.2409.12122} {Qwen2.5-math technical report: Toward mathematical expert model via self-improvement}.
\newblock \emph{CoRR}, abs/2409.12122.

\bibitem[{Ye et~al.(2025)Ye, Huang, Xiao, Chern, Xia, and Liu}]{1.16_data_limo}
Yixin Ye, Zhen Huang, Yang Xiao, Ethan Chern, Shijie Xia, and Pengfei Liu. 2025.
\newblock \href {https://doi.org/10.48550/ARXIV.2502.03387} {{LIMO:} less is more for reasoning}.
\newblock \emph{CoRR}, abs/2502.03387.

\bibitem[{Yu et~al.(2025)Yu, Zhang, Zhu, Yuan, Zuo, Yue, Fan, Liu, Liu, Liu, Lin, Lin, Ma, Sheng, Tong, Zhang, Zhang, Zhang, Zhu, Zhu, Chen, Chen, Wang, Yu, Dai, Song, Wei, Zhou, Liu, Ma, Zhang, Yan, Qiao, Wu, and Wang}]{1.10_dapo}
Qiying Yu, Zheng Zhang, Ruofei Zhu, Yufeng Yuan, Xiaochen Zuo, Yu~Yue, Tiantian Fan, Gaohong Liu, Lingjun Liu, Xin Liu, Haibin Lin, Zhiqi Lin, Bole Ma, Guangming Sheng, Yuxuan Tong, Chi Zhang, Mofan Zhang, Wang Zhang, Hang Zhu, and 16 others. 2025.
\newblock \href {https://doi.org/10.48550/ARXIV.2503.14476} {{DAPO:} an open-source {LLM} reinforcement learning system at scale}.
\newblock \emph{CoRR}, abs/2503.14476.

\bibitem[{Yuan et~al.(2025{\natexlab{a}})Yuan, Yue, Zhu, Fan, and Yan}]{1.11_RLVR}
Yufeng Yuan, Yu~Yue, Ruofei Zhu, Tiantian Fan, and Lin Yan. 2025{\natexlab{a}}.
\newblock \href {https://doi.org/10.48550/ARXIV.2503.01491} {What's behind ppo's collapse in long-cot? value optimization holds the secret}.
\newblock \emph{CoRR}, abs/2503.01491.

\bibitem[{Yuan et~al.(2025{\natexlab{b}})Yuan, Yue, Zhu, Fan, and Yan}]{1.19_vc_ppo}
Yufeng Yuan, Yu~Yue, Ruofei Zhu, Tiantian Fan, and Lin Yan. 2025{\natexlab{b}}.
\newblock \href {https://doi.org/10.48550/ARXIV.2503.01491} {What's behind ppo's collapse in long-cot? value optimization holds the secret}.
\newblock \emph{CoRR}, abs/2503.01491.

\bibitem[{Yue et~al.(2025)Yue, Yuan, Yu, Zuo, Zhu, Xu, Chen, Wang, Fan, Du, Wei, Yu, Liu, Liu, Liu, Lin, Lin, Ma, Zhang, Zhang, Zhang, Zhu, Zhang, Liu, Wang, Wu, and Yan}]{2.5_vapo_rlvr}
Yu~Yue, Yufeng Yuan, Qiying Yu, Xiaochen Zuo, Ruofei Zhu, Wenyuan Xu, Jiaze Chen, Chengyi Wang, TianTian Fan, Zhengyin Du, Xiangpeng Wei, Xiangyu Yu, Gaohong Liu, Juncai Liu, Lingjun Liu, Haibin Lin, Zhiqi Lin, Bole Ma, Chi Zhang, and 8 others. 2025.
\newblock \href {https://arxiv.org/abs/2504.05118} {Vapo: Efficient and reliable reinforcement learning for advanced reasoning tasks}.
\newblock \emph{Preprint}, arXiv:2504.05118.

\bibitem[{Zeng et~al.(2025{\natexlab{a}})Zeng, Huang, Liu, Liu, He, Ma, and He}]{2.2_rlvr}
Weihao Zeng, Yuzhen Huang, Qian Liu, Wei Liu, Keqing He, Zejun Ma, and Junxian He. 2025{\natexlab{a}}.
\newblock \href {https://doi.org/10.48550/ARXIV.2503.18892} {Simplerl-zoo: Investigating and taming zero reinforcement learning for open base models in the wild}.
\newblock \emph{CoRR}, abs/2503.18892.

\bibitem[{Zeng et~al.(2025{\natexlab{b}})Zeng, Huang, Liu, Liu, He, Ma, and He}]{4_9_simplerl}
Weihao Zeng, Yuzhen Huang, Qian Liu, Wei Liu, Keqing He, Zejun Ma, and Junxian He. 2025{\natexlab{b}}.
\newblock \href {https://doi.org/10.48550/ARXIV.2503.18892} {Simplerl-zoo: Investigating and taming zero reinforcement learning for open base models in the wild}.
\newblock \emph{CoRR}, abs/2503.18892.

\bibitem[{Zha et~al.(2025)Zha, Bhat, Lai, Yang, Jiang, Zhong, and Hu}]{3_1.4_survey}
Daochen Zha, Zaid~Pervaiz Bhat, Kwei{-}Herng Lai, Fan Yang, Zhimeng Jiang, Shaochen Zhong, and Xia~Ben Hu. 2025.
\newblock \href {https://doi.org/10.1145/3711118} {Data-centric artificial intelligence: {A} survey}.
\newblock \emph{{ACM} Comput. Surv.}, 57(5):129:1--129:42.

\bibitem[{Zhang et~al.(2024)Zhang, Zhoubian, Hu, Yue, Dong, and Tang}]{3_1.2_mcts_sft}
Dan Zhang, Sining Zhoubian, Ziniu Hu, Yisong Yue, Yuxiao Dong, and Jie Tang. 2024.
\newblock \href {http://papers.nips.cc/paper\_files/paper/2024/hash/76ec4dc30e9faaf0e4b6093eaa377218-Abstract-Conference.html} {Rest-mcts*: {LLM} self-training via process reward guided tree search}.
\newblock In \emph{Advances in Neural Information Processing Systems 38: Annual Conference on Neural Information Processing Systems 2024, NeurIPS 2024, Vancouver, BC, Canada, December 10 - 15, 2024}.

\bibitem[{Zhang et~al.(2025)Zhang, Wang, Cheng, Zhuang, Lin, Zhang, Wang, Cui, Wang, Peng et~al.}]{1.7_SRPO}
Xiaojiang Zhang, Jinghui Wang, Zifei Cheng, Wenhao Zhuang, Zheng Lin, Minglei Zhang, Shaojie Wang, Yinghan Cui, Chao Wang, Junyi Peng, and 1 others. 2025.
\newblock \href {https://arxiv.org/abs/2504.14286} {Srpo: A cross-domain implementation of large-scale reinforcement learning on llm}.
\newblock \emph{arXiv preprint arXiv:2504.14286}.

\bibitem[{Zhou et~al.(2022)Zhou, Wu, Zhu, and Liang}]{3_1.9_sample_weighting}
Xiaoling Zhou, Ou~Wu, Weiyao Zhu, and Ziyang Liang. 2022.
\newblock \href {https://doi.org/10.1007/978-3-031-26409-2\_5} {Understanding difficulty-based sample weighting with a universal difficulty measure}.
\newblock In \emph{Machine Learning and Knowledge Discovery in Databases - European Conference, {ECML} {PKDD} 2022, Grenoble, France, September 19-23, 2022, Proceedings, Part {III}}, volume 13715 of \emph{Lecture Notes in Computer Science}, pages 68--84. Springer.

\end{thebibliography}
\newpage
\appendix

% \section{Integrated Optimization Pipeline with PG-Sampling and LP-Weighting} 
% The Integrated Training Process with PG-Sampling and LP-Weighting are detailed in \ref{sec:Integrated Training Process with PG-Sampling and LP-Weighting}.
% \input{Fig/appendix/algo}

\section{Integrated Pipeline with Our LPPO}
\label{sec:integrated-training}

This section details the integrated progressive opptimization RLVR pipeline that combines PG-Sampling and LP-Weighting to improve sample efficiency and training dynamics. The full algorithm is summarized in Algorithm~\ref{sec:Integrated Training Process with PG-Sampling and LP-Weighting}.
\label{sec:Integrated Training Process with PG-Sampling and LP-Weighting}
\begin{algorithm*}[t]
\small
\caption{Integrated Training Process with Proposed LPPO}
\label{alg:rlvr_pg_lp}
\begin{algorithmic}[1]
\Require Standard RLVR dataset $\mathcal{D}$, PG-sampling subset $\mathcal{D}_{pg}$, initial policy parameters $\theta$, EMA decay $\alpha$, LP-weighting sensitivity $\kappa$, LP-weighting bias $b$
\Ensure Updated policy parameters $\theta$
\State \textit{// 1. Initialize EMA of pass rates}
\State Initialize EMA of pass rates $\overline{p_i}(0)\gets 0$ for all samples $i$
\For{epoch $t=1$ to $T$}
    \State \textit{// 2. Compute pass rates}
    \For{each sample $i$ in $\mathcal{D}\cup \mathcal{D}_{pg}$}
        \State Roll out current policy $\pi_\theta$ on sample $i$ to compute $\mathrm{pass\_rate}_i(t)$
        \State Compute reward $r_i$ for trajectories from sample $i$
        \If{$0 < \mathrm{pass\_rate}_i(t) < 1$}
            \State Add these trajectories (with reward $r_i$) to the $batch$ for policy update
        \EndIf
    \EndFor
    \State \textit{// 3. PG-Sampling for difficult samples}
    \For{each sample $i$ in $\mathcal{D}_{pg}$ with $\mathrm{pass\_rate}_i(t)=0$}
        \State Generate a prefix $S_{\text{pre},i}$ for sample $i$ according to Eq.~\ref{eq:prefix_length} and Eq.~\ref{eq:prefix_definition}
        \State Generate the remaining sequence (suffix) $S_{\text{rem},i}$ for this prefix using Eq.~\ref{eq:remainder_generation}
        \State Compute reward $r_i$ for trajectories from sample $i$
        \State Add these prefix-guided trajectories ($S_{\text{rem},i}$, with reward $r_i$) to the $batch$ for policy update
    \EndFor
    \State \textit{// 4. Update pass-rate statistics and LP weights}
    \For{each sample $i$ in $batch$} 
        \State Update EMA of pass rate: $\overline{p_i}(t)\gets (1-\alpha)\overline{p_i}(t-1) + \alpha\,\mathrm{pass\_rate}_i(t)$ \Comment{(Corresponds to Eq.~\ref{eq:ema_pass_rate})}
        \State Compute learning progress: $\Delta_i(t)\gets \overline{p_i}(t)-\overline{p_i}(t-1)$ \Comment{(Corresponds to Eq.~\ref{eq:learning_progress_delta})}
        \State Compute LP-weight: $w_i(t)\gets \sigma(\kappa \cdot \Delta_i(t)) + b$ \Comment{(Corresponds to Eq.~\ref{eq:dynamic_weight}, where $\sigma$ is the sigmoid function)}
    \EndFor
    \State \textit{// 5. Compute weighted advantages}
    \For{each sample $i$ and rollout index $k$}
        \State Weight the advantage: $\hat{A}'_{i,k}\gets w_i(t)\,\hat{A}_{i,k}$ \Comment{(Corresponds to Eq.~\ref{eq:weighted_advantage})}
    \EndFor
    \State \textit{// 6. Policy update}
    \State Update policy $\pi_\theta$ via GRPO \Comment{(Corresponds to Eq.~\ref{eq:lp_grpo_objective_no_kl})}
    \State \textit{// 7. Online data curation}
    \State Remove (or skip) consistently solved samples (with $\mathrm{pass\_rate}_i(t)=1.0$) in $\mathcal{D}\cup \mathcal{D}_{pg}$ from training
\EndFor
\end{algorithmic}
\end{algorithm*}

\section{Additional Experiment Settings}
\label{sec:AdditionalExperimentSettings}
\paragraph{Prompt}
We use the same template as Qwen-MATH \footnote{https://huggingface.co/Qwen/Qwen2.5-Math-7B-Instruct}, shown in Fig.~\ref{fig:Prompt_template}, for our training and evaluation. For PG-sampling, which combines the solution and the problem for difficult questions, we inject the prefix \texttt{solution\_prefix} into the prompt for training, as shown in Fig.~\ref{fig:prefix_Prompt_template}.

\begin{figure}[t]
\centering
\begin{tcolorbox}[
  colback=white, 
  colframe=gray,
  coltitle=white,
  colbacktitle=gray,
  title=\textbf{Prompt},
  fonttitle=\bfseries,
  arc=2mm,
  boxrule=0.8pt,
  left=4mm,
  right=4mm,
  top=2mm,
  bottom=2mm
]
{\color{blue}\verb|system|}

\verb|Please reason step by step, and put |

\verb|your final answer within \\boxed{}.|

{\color{blue}\verb|user|}

{\color{red}\verb|{question}|}

{\color{blue}\verb|assistant|}

{\color{red}\verb|{answer}|}
\end{tcolorbox}
\caption{Prompt template.}
\label{fig:Prompt_template}
\end{figure}

\begin{figure}[t]
\centering
\begin{tcolorbox}[
  colback=white, 
  colframe=gray,
  coltitle=white,
  colbacktitle=gray,
  title=\textbf{Prompt For PG-Sampling},
  fonttitle=\bfseries,
  arc=2mm,
  boxrule=0.8pt,
  left=4mm,
  right=4mm,
  top=2mm,
  bottom=2mm
]
{\color{blue}\verb|system|}

\verb|Please reason step by step, and put |

\verb|your final answer within \\boxed{}.|

{\color{blue}\verb|user|}

{\color{red}\verb|{question}|}

{\color{blue}\verb|assistant|}

{\color{red}\verb|{solution_prefix} + {answer}|}
\end{tcolorbox}
\caption{Prompt template for PG-Sampling.}
\label{fig:prefix_Prompt_template}
\end{figure}

\paragraph{Reward Design}
We focus exclusively on the model’s reasoning ability, not on exposing its entire chain of thought; hence we do not adopt the format reward for Deepseek-R1 ~\cite{1.1_deepseek_r1}.  
To minimise the risk of reward-hacking that can arise from elaborate scoring schemes, we employ a deliberately minimal, rule-based metric.

The sole component is mathematical correctness, denoted \(R_{\text{math}}\); no extra credit is awarded for formatting or auxiliary details.  
After the model produces its final answer, an automatic verifier checks equivalence with the ground truth:

\[
R_{\text{math}} =
\begin{cases}
1, & \text{if the answer is completely correct}, \\[4pt]
0, & \text{otherwise}.
\end{cases}
\]

Accordingly, the total reward is simply

\[
R = R_{\text{math}}.
\]

\begin{figure}[ht]
    \centering
    \includegraphics[width=0.45\textwidth]{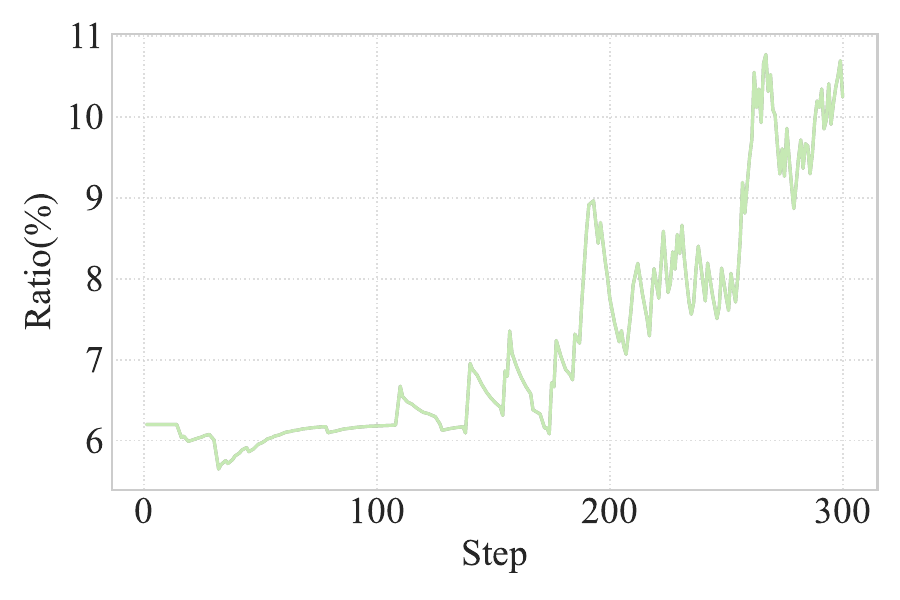} % 替换路径
    \caption{Ratio of PG-Sampling augmented samples over training steps.}
    \label{fig:prefix_ratio}
\end{figure}

\section{Further Exploration on PG-Sampling}

\paragraph{Ratio of PG-Sampling Augmented Samples} Figure~\ref{fig:prefix_ratio} shows the ratio of PG-Sampling augmented samples during training. Due to the presence of online data curation, the proportion of samples affected by the PG-Sampling strategy gradually increases as the training proceeds, rising steadily from an initial value of around 6\% and eventually reaching approximately 11\% at step 300. This gradual growth occurs because online data curation continuously filters out samples that become too easy, leaving behind those challenging instances that trigger the PG-Sampling augmentation. As a result, the proportion of prefix-guided samples within each training batch gradually increases over time.

% 由于PG-sampling策略采用了来自专家模型的解答前缀作为模型输入的一部分，在进行该训练时，模型会在已知部分解题逻辑和专家题解风格的情况下进行推理。在这种情况下，模型是否会在无SFT损失的情况下，通过模仿专家解答的逻辑？结论是该训练方式并不会令模型的输出更加接近专家输出。

\begin{figure*}[h]
    \centering
    \includegraphics[width=0.8\textwidth]{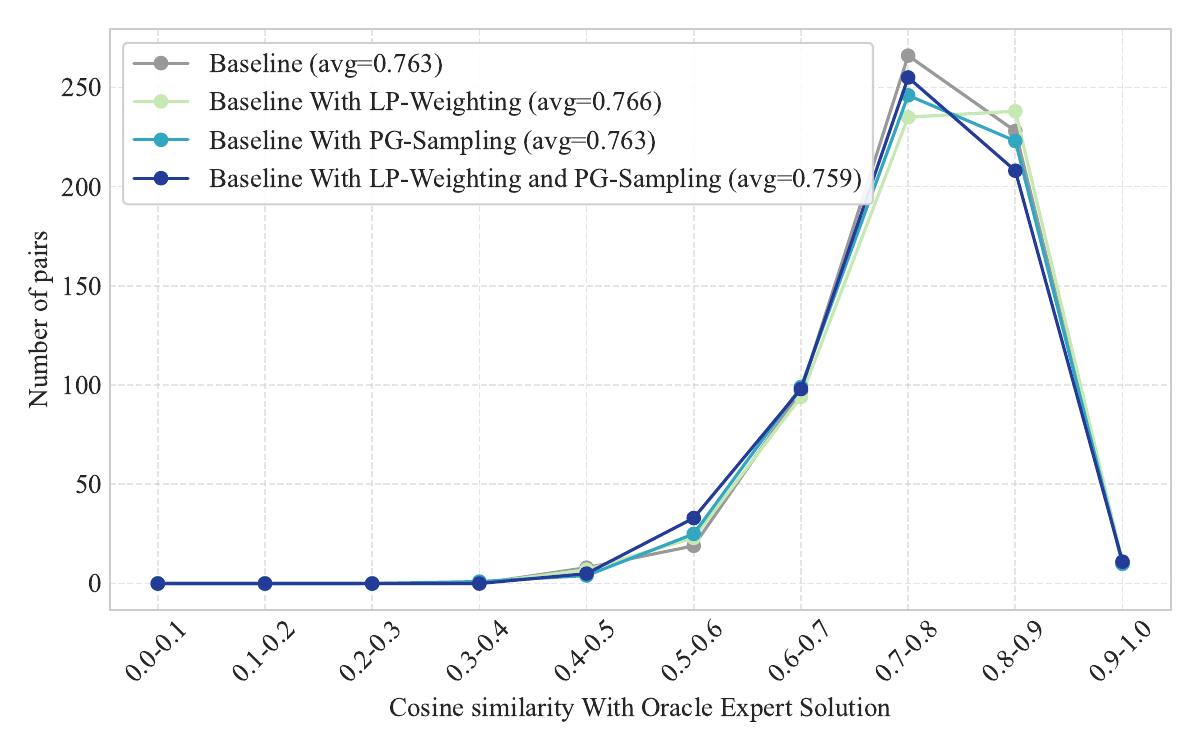} % 替换路径
    \caption{The Similarity between Oracle Expert Solution with/without PG-Sampling.}
    \label{fig:sim_pg_sampling}
\end{figure*}

\begin{table*}[t]
  \centering
  \small
  \setlength{\tabcolsep}{5pt}
  \renewcommand{\arraystretch}{1.15}
\begin{tabular}{lccccccc}
\hline
\textbf{Setting} & AIME24        & AIME25        & AMC23         & MATH-500 & Minerva       & Olympiad      & Avg.          \\ \hline
\multicolumn{8}{c}{\textit{Larger backbone: Qwen-2.5-14B}}                                                                                  \\ \hline
Baseline (GRPO)  & \textbf{13.3}          & 13.3          & 57.5          & 79.9                     & \textbf{47.3}          & 44.7          & 42.7          \\
+\,LPPO          & \textbf{13.3} & \textbf{20.0} & \textbf{62.5} & \textbf{82.2}            & 46.0 & \textbf{46.1} & \textbf{45.0} \\ \hline
\multicolumn{8}{c}{\textit{Different backbone: Llama-3.2-3B-Instruct}}                                                                        \\ \hline
Baseline (GRPO)  & 16.7          & 3.3           & 25.8          & \textbf{58.0}                     & 24.9          & \textbf{23.1}          & 25.3          \\
+\,LPPO          & \textbf{20.0} & \textbf{6.7}  & \textbf{35.0} & 57.8            & \textbf{25.0} & 23.0 & \textbf{27.9} \\ \hline
\multicolumn{8}{c}{\textit{Different learner: Qwen-2.5-Math-7B + REINFORCE++}}                                                              \\ \hline
Baseline         & 26.7          & 10.0          & 60.8          & \textbf{81.2}                     & 45.3          & 43.7          & 44.6          \\
+\,LPPO          & \textbf{43.3} & \textbf{13.3} & \textbf{65.0} & 81.0            & \textbf{45.5} & \textbf{44.0} & \textbf{48.7} \\ \hline
\end{tabular}
  \caption{Pass@1 accuracy (\%) of LPPO across diverse scenarios. LPPO consistently brings +2–4 pp absolute improvements over each corresponding baseline without any hyper-parameter retuning.}
  \label{tab:lppo_diverse}
\end{table*}

\paragraph{Does PG-Sampling Induce Expert-Like Reasoning?} The PG-Sampling strategy incorporates expert-generated partial solutions as inputs, potentially prompting the model to imitate the expert's reasoning pattern even without explicit SFT loss. To examine whether this sampling technique leads to outputs that closely align with expert solutions, we use the all-MiniLM-L6-v2 model \cite{4_12_sentence_transformer} to measure the semantic similarity between the model-generated and oracle expert solutions.

Figure~\ref{fig:sim_pg_sampling} illustrates the cosine similarity between the embeddings of model-generated solutions and oracle expert solutions, comparing training with and without PG-Sampling. Notably, the similarity curves for both methods exhibit similar trends and convergence behaviors. This observation indicates that incorporating expert solution prefixes does not inherently make the model's generated outputs significantly closer to expert reasoning styles. Consequently, the primary benefit of PG-Sampling is enhancing exploration efficiency rather than promoting imitation of expert solutions.

\section{LPPO under Diverse Scenarios}
\label{app:appendix_diverse}
This appendix complements Section~\ref{sec:lppo_diverse} in the main text; the complete numerical results are listed in Table~\ref{tab:lppo_diverse} for easy cross-reference.

\end{document}